\documentclass{article}

\PassOptionsToPackage{round}{natbib}

\usepackage[final]{neurips_2025}

\usepackage{definitions}

\usepackage[utf8]{inputenc} 
\usepackage[T1]{fontenc}    
\usepackage{url}            
\usepackage{booktabs}       
\usepackage{amsfonts}       
\usepackage{nicefrac}       
\usepackage{microtype}      
\usepackage{xcolor}         
\usepackage{tcolorbox}
\usepackage{subcaption}
\usepackage{pifont}
\usepackage{adjustbox}
\usepackage{graphicx}
\usepackage{colortbl}
\usepackage{geometry}
\usepackage{enumitem}
\usepackage{float}

\definecolor{electriclavender}{rgb}{0.96, 0.73, 1.0}
\definecolor{darkblue}{HTML}{1A254B}
\definecolor{lightblue}{HTML}{A7BED3}
\definecolor{blue}{HTML}{114083}
\definecolor{green}{HTML}{71D49F}
\definecolor{pink}{HTML}{F2545B}
\definecolor{red}{HTML}{F2545B}
\definecolor{lightgray}{HTML}{C3BABA}
\definecolor{darkgray}{HTML}{9A8F97}

\usepackage[hidelinks]{hyperref}
\hypersetup{
    colorlinks=true,
    linkcolor=blue,
    urlcolor=blue,
    citecolor=blue,
    anchorcolor=blue}
\usepackage{url}

\newcommand{\cmark}{{\color{green}\ding{51}}}
\newcommand{\xmark}{{\color{red}\ding{55}}}

\usepackage{newfloat}
\DeclareFloatingEnvironment[
    fileext=loe,
    listname={List of Examples},
    name=Example,
    placement=tbhp,
]{exmpl}

\title{\our: A Multimodal Sequential Clinical Decision-Making Benchmark in Oncology}

\author{
\textbf{Kiril Vasilev}$^{1}$\thanks{These authors contributed equally.} \qquad
\textbf{Alexandre Misrahi}$^{2*}$ \qquad
\textbf{Eeshaan Jain}$^{2*}$\thanks{Correspondence to: \texttt{eeshaan.jain@epfl.ch}} \\
\textbf{Phil Cheng}$^{3}$ \qquad
\textbf{Petros Liakopoulos}$^{3}$ \qquad
\textbf{Olivier Michielin}$^{3}$ \\
\textbf{Michael Moor}$^{1\ddagger}$ \qquad
\textbf{Charlotte Bunne}$^{2}$\thanks{Co-last authorship.} \\[3pt]
$^{1}$ETH Zürich \quad $^{2}$EPFL \quad $^{3}$HUG \\[6pt]
\raisebox{-0.3\height}{\includegraphics[width=0.035\textwidth]{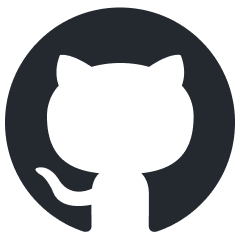}}
\href{https://github.com/bunnelab/MTBBench}{\texttt{github.com/bunnelab/MTBBench}} \\
\raisebox{-0.3\height}{\includegraphics[width=0.035\textwidth]{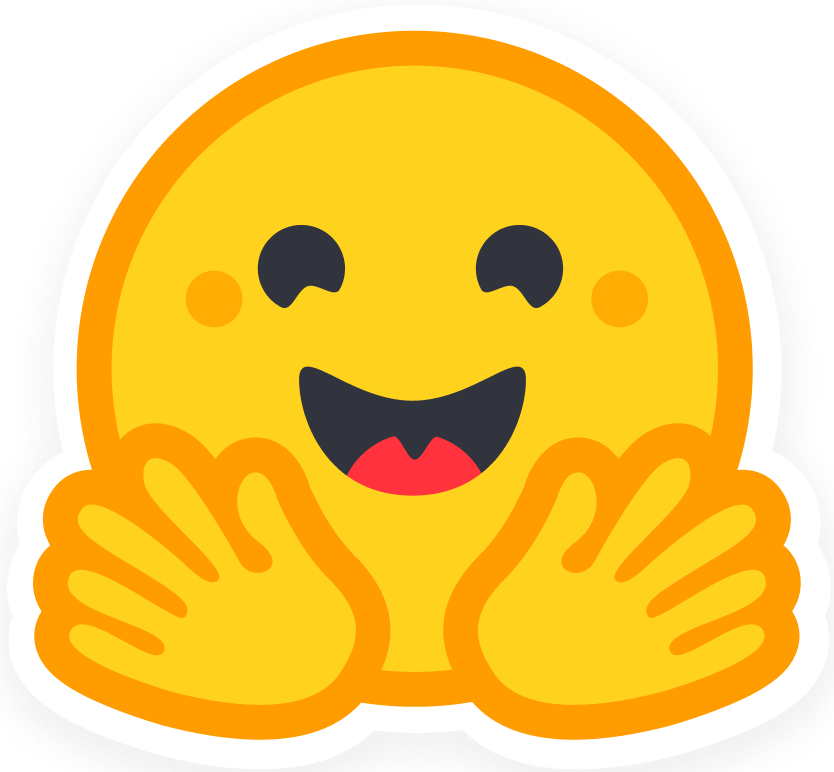}}
\href{https://huggingface.co/datasets/EeshaanJain/MTBBench}{\texttt{huggingface.co/datasets/EeshaanJain/MTBBench}}
}

\begin{document}

\maketitle

\begin{abstract}
    Multimodal Large Language Models (LLMs) hold promise for biomedical reasoning, but current benchmarks fail to capture the complexity of real-world clinical workflows. Existing evaluations primarily assess unimodal, decontextualized question-answering, overlooking multi-agent decision-making environments such as Molecular Tumor Boards (MTBs). MTBs bring together diverse experts in oncology, where diagnostic and prognostic tasks require integrating heterogeneous data and evolving insights over time. Current benchmarks lack this longitudinal and multimodal complexity.
    We introduce \textbf{\our}, an agentic benchmark simulating MTB-style decision-making through clinically challenging, multimodal, and longitudinal oncology questions. Ground truth annotations are validated by clinicians via a co-developed app, ensuring clinical relevance. We benchmark multiple open and closed-source LLMs and show that, even at scale, they lack reliability---frequently hallucinating, struggling with reasoning from time-resolved data, and failing to reconcile conflicting evidence or different modalities. To address these limitations, \our goes beyond benchmarking by providing an agentic framework with foundation model-based tools that enhance multi-modal and longitudinal reasoning, leading to task-level performance gains of up to 9.0\% and 11.2\%, respectively.
    Overall, \our offers a challenging and realistic testbed for advancing multimodal LLM reasoning, reliability, and tool-use with a focus on MTB environments in precision oncology.

\end{abstract}

\vspace{-10pt}
\section{Introduction}

Recent advances in large multi-modal and language models have opened the door to general-purpose clinical agents capable of reasoning across diverse biomedical tasks \citep{moor2023foundation}. Vision-language models can describe pathology images \citep{Lu2024, Dai2025, CONCH}, LLMs can summarize clinical notes \citep{Choudhuri2025.01.19.25320797, yang2024clinicalmambagenerativeclinicallanguage}, and medical agents are increasingly able to query tools, retrieve knowledge, and even hold multi-turn clinical conversations \citep{schmidgall2024agentclinicmultimodalagentbenchmark, wang2025surveyllmbasedagentsmedicine}. These developments have prompted growing interest in using agents to support complex workflows \citep{wang2024perspectiveadaptinggeneralistai, wang2025txgemmaefficientagenticllms, gao2024empoweringbiomedicaldiscoveryai, Lee2024, yue2024clinicalagentclinicaltrialmultiagent, fallahpour2025medraxmedicalreasoningagent} like those seen in \textit{molecular tumor boards} (MTBs) \citep{Tsimberidou2023}, where oncologists, radiologists, pathologists, and geneticists jointly analyze a patient’s evolving case (Fig.~\ref{fig:mtb}).

\begin{figure}[t]
    \centering
    \includegraphics[width=\linewidth]{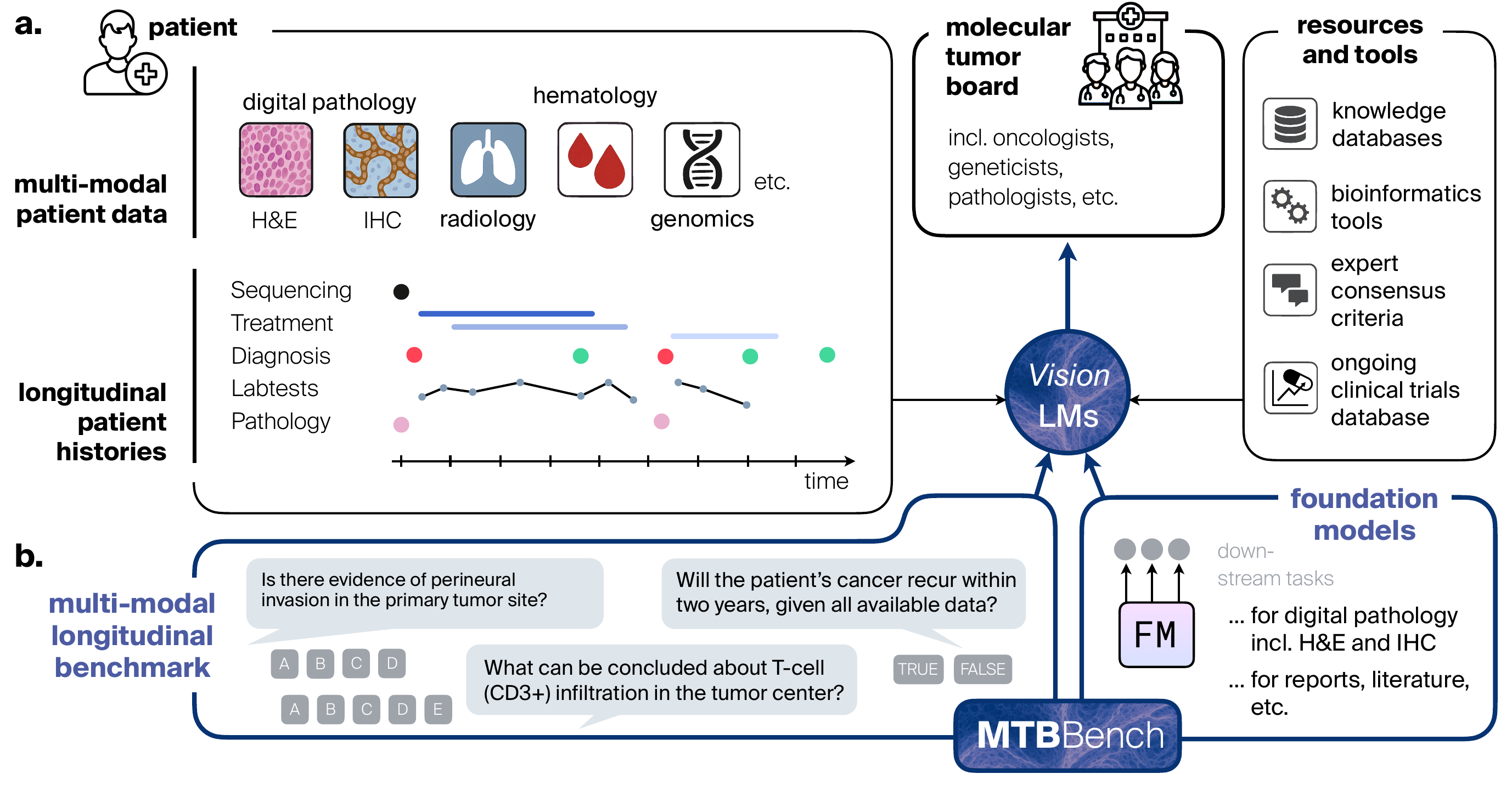}
    \caption{\footnotesize \textbf{The \our benchmark and agent framework.} \textbf{a.} \our simulates molecular tumor board workflows, presenting agents with longitudinal, multi-modal patient data (H\&E, IHC, hematology, and genomics) along with temporally distributed clinical events. Agents are tasked with integrating this information to support complex decision-making. \textbf{b.} \our allows benchmarking agents on their ability to reason across modalities and time in order to accurately tackle clinical questions concerning diagnosis, prognosis, and biomarker interpretation. Lastly, we introduce an agentic framework that enables querying both external tools and pretrained foundation models, allowing agents to more effectively reason over complex, multi-modal and temporally resolved clinical information.}
    \label{fig:overview}
\end{figure}

However, the evaluation of such agents remains underdeveloped. Existing benchmarks typically frame tasks as \textit{static, uni-modal, single-turn question-answering problems}, where the model is given all necessary inputs at once and evaluated on its ability to predict a discrete answer. This setup diverges sharply from how clinical decisions are made in practice. Real-world oncology reasoning is \textit{interactive, temporal, and multi-modal}: physicians accumulate information over time, integrate findings from multiple data types (e.g., hematoxylin and eosin (H\&E) staining, immunohistochemistry (IHC) staining, radiology, blood values, genomics), and make provisional decisions that are updated as new evidence emerges (Fig.~\ref{fig:overview}a). To be useful in these settings, AI agents must not only understand each modality, but also \textit{query, contextualize, and reconcile} information across modalities and time---capabilities rarely assessed in current evaluations.

Recent works such as \texttt{MedAgentBench} \citep{jiang2025medagentbenchrealisticvirtualehr}, \texttt{MediQ} \citep{li2024mediqquestionaskingllmsbenchmark}, and \texttt{MedJourney} \citep{medjourney} take steps toward interactive or longitudinal evaluation, but typically in limited or uni-modal contexts (e.g., textual EHRs) \citep{kweon2024ehrnoteqa} (Table~\ref{tab:comparison}). Likewise, emerging studies on multi-modal agents demonstrate strong promise but lack standardized evaluation across \textit{longitudinal patient trajectories} \citep{li2024mediqquestionaskingllmsbenchmark}. Most importantly, these agents are not tested under the cognitive demands of tasks that mirror MTB decision-making: questions involving partial data, sequential updates, conflicting information, highly heterogeneous and different modalities, and high-stakes outcomes.

To address this gap, we introduce \textbf{\our}, an oncology-focused benchmark for evaluating AI agents in \textit{longitudinal, multi-modal clinical reasoning}. Inspired by the structure and decision flow of real molecular tumor boards, \our simulates patient case reviews where agents must process heterogeneous patient data across time---including pathology slides, lab data, pathological, surgical and genomic reports---and answer clinically meaningful questions at each step. Questions span diverse task types, including diagnostic classification, spatial biomarker interpretation, and outcome, progression, or recurrence prediction (Fig.~\ref{fig:overview}b). Importantly, the benchmark is \textit{validated by clinicians} using a custom-built expert annotation platform (Fig.~\ref{fig:comanion_app}, for details see Appendix~\ref{sec:companion_app}), ensuring both the clinical plausibility of the data and the correctness of model evaluation.

Beyond benchmark construction, we also introduce a modular \textit{agentic framework} designed to interface with different tools as well as \textbf{pretrained foundation models} (Fig.~\ref{fig:overview}b). These include models trained on large-scale digital pathology datasets, reports, literature, and other domain-specific modalities. Agents can query these foundation models as part of their reasoning process---invoking them when needed to interpret image regions, extract genomic signatures, or cross-reference trial data—thus mirroring how expert clinicians rely on specialized resources in practice. This framework enables flexible, multi-step decision-making and substantially enhances the agent’s ability to synthesize information across modalities and time.

Concretely, our \textbf{main contributions} are:
\begin{enumerate}[align=left, label=\roman*., leftmargin=*, noitemsep]
\itemsep0em
    \item \textbf{A benchmark for longitudinal, multi-modal clinical reasoning.} \our simulates MTB-style decision-making with temporally evolving patient data across modalities—H\&E, IHC, hematology, and genomics—and includes complex, expert-curated questions reflecting real tumor board workflows.
    \item \textbf{Systematic evaluation of vision-language models.} We benchmark diverse open and closed-source models and find that performance improves with exposure to more modalities, emphasizing the importance of integrated multi-modal context.
    \item \textbf{An agent framework for tool and foundation model integration.} Our agent framework enables dynamic access to external tools and pretrained modality-specific foundation models, significantly boosting reasoning accuracy across tasks.
    \item \textbf{Expert-validated data and reproducible tools.} We release curated benchmark data, expert-reviewed annotations, agent logs, and tools to support rigorous and reproducible evaluation of clinical agents.
\end{enumerate}

\begin{table}[t]
\centering
\caption{Comparison of \our with existing clinical and biomedical benchmarks.}
\label{tab:comparison}
\begin{adjustbox}{max width=\textwidth}
\begin{tabular}{lccccc}
\toprule
\textbf{Benchmark} & \textbf{Multi-Modal} & \textbf{Longitudinal} & \textbf{Interactive / Multi-Agent} & \textbf{Clinician-Annotated} & \textbf{Domain} \\
\midrule
\texttt{MC-BEC} \citep{chen_multimodal_2023} & \cmark & \cmark & \xmark & \xmark & Emergency Medicine \\
\texttt{Asclepius} \citep{wang2024asclepius} & \cmark & \xmark & \xmark & \xmark & General Medicine \\
\texttt{MedJourney} \citep{medjourney} & \xmark & \cmark & \xmark & \xmark & General Medicine \\
\texttt{EHRNoteQA} \citep{kweon2024ehrnoteqa} & \xmark & \cmark & \xmark & \cmark & General Medicine \\
\texttt{MediQ} \citep{li2024mediqquestionaskingllmsbenchmark} & \xmark & \xmark & \cmark & \xmark & General Medicine \\
\texttt{ClinicBench} \citep{Liu2024ClinicBench} & \xmark & \cmark & \xmark & \xmark & General Medicine \\
\texttt{HEST-1k} \citep{jaume2024hest} & \cmark & \xmark & \xmark & \cmark & Pathology \\
\texttt{BixBench} \citep{mitchener2025bixbenchcomprehensivebenchmarkllmbased} & \xmark & \xmark & \xmark & \xmark & Bioinformatics \\
\addlinespace
\textbf{\our} (ours) & \textbf{\cmark} & \textbf{\cmark} & \textbf{\cmark} & \textbf{\cmark} & \textbf{Precision Oncology} \\
\bottomrule
\end{tabular}
\end{adjustbox}
\end{table}

\vspace{-5pt}
\section{\our: A Multimodal Sequential Clinical Decision-Making Benchmark in Oncology} \label{sec:benchmark}
\vspace{-5pt}

\subsection{Motivation and Positioning}
LLMs have shown increasing promise across medical domains, but current benchmarks remain disconnected from the realities of clinical workflows. \citet{ferber2024autonomousartificialintelligenceagents} introduce an agent for clinical decision making, however it is limited to general tool frameworks and to single-image inputs. Others evaluate unimodal, static question-answering tasks without requiring the model to gather information, reconcile conflicting inputs, or reason over time \citep{li2025agenthospitalsimulacrumhospital}. In contrast, real clinical decision-making—especially in molecular tumor boards---is inherently multimodal, interactive, and longitudinal.

\our is a benchmark designed to close this gap. It evaluates how well AI agents can simulate an MTB-style setting, where patient cases evolve across time and require integration of imaging, lab, pathology, genomic, and textual information. \our captures three essential dimensions lacking in prior work:
\begin{itemize}[leftmargin=*, noitemsep]
\itemsep0em
    \item[] \textbf{Multimodality}: Clinical data spans digital pathology (H\&E, IHC), hematology, radiology, and genomics.
    \item[] \textbf{Longitudinality}: Patient histories unfold over multiple timepoints, with temporally ordered updates.
    \item[] \textbf{Agent workflow}: Models must actively request relevant information, access tools, and answer in multi-turn settings.
\end{itemize}

As shown in Table~\ref{tab:comparison}, \our is the first benchmark to jointly address these three dimensions. All questions are expert-validated through a co-developed application, which we further describe below. Together, these components establish \our as a framework for evaluating the capabilities of AI agents in clinical settings that require multi-modal and longitudinal understanding. For further related works, see Appendix~\ref{sec:related_works}.

\subsection{A Benchmark for Molecular Tumor Boards}
\our comprises both multiple-choice and true/false questions designed to be answered within a clinically grounded, agent-based framework. Questions span multimodal and longitudinal patient data, requiring the agent to retrieve, interpret, and reason over diverse evidence sources under realistic constraints that mirror the sequential and evolving nature of clinical decision-making.

\paragraph{Companion application for expert validation} 
To support expert review of the benchmark, we developed a web-based interface allowing clinicians to inspect the clinical context, browse pathology and IHC images, and annotate feedback for each question. The interface presents structured case descriptions alongside image thumbnails of different modalities that open full-resolution slides on demand (Fig.~\ref{fig:comanion_app}). Clinicians can view grouped images by region and marker (e.g., CD3, CD163) and provide detailed assessments directly linked to individual Q\&A items. This tool enabled efficient, structured validation of questions and answers by domain experts.

\subsubsection{\our-Multimodal}

\paragraph{Dataset} We curated a subset of 26 patient cases from the \hc dataset (CC BY 4.0) \citep{hancock}, a multimodal repository of head and neck cancer patients that includes demographic, pathological, hematological, surgical, and histological data. For each selected patient, an average of 40 modality-specific files are available, including 1.2 H\&E slides, 26.2 IHC images, and one hematology report. Among these cases, 32 include a primary tumor H\&E slide, and 17 contain at least one lymph node slide. Notably, 27 of the primary tumor slides are accompanied by annotated regions of interest.

\paragraph{Q\&A design} We generate 390 multimodal question-answer pairs (15 per patient) using GPT-4o, with select questions verified through expert-in-the-loop review through the companion app (Fig.~\ref{fig:comanion_app}, Appendix~\ref{sec:companion_app}). The questions span a range of modalities and clinical reasoning tasks: 2 H\&E-based, 4 IHC-based, 3 joint H\&E+IHC, 4 hematology-based, and one question each related to clinical outcome and cancer recurrence.

\paragraph{Tasks} 
The multi-modal track unfolds in three clinically grounded stages. The first focuses on pathological image interpretation (e.g., of H\&E, IHC images): agents must infer histologic subtypes and assess spatial patterns of immune infiltration (e.g., CD3\textsuperscript{+}, CD8\textsuperscript{+} T cells, CD68\textsuperscript{+}, CD163\textsuperscript{+} macrophages) across tumor subregions such as the invasion front and tumor center. Intermediate questions probe immune correlates of pathology, such as whether lymphovascular or perineural invasion associates with distinct immune profiles. The second stage evaluates hematologic reasoning in a preoperative context—models analyze lab parameters (e.g., CRP, MPV, leukocyte subtypes, creatinine) to infer infection risk, bleeding tendency, renal impairment, and thromboembolic predisposition. Finally, in the post-surgical stage, agents must integrate pathology and lab findings to predict high-level outcomes, including 5-year survival and 2-year recurrence, simulating tumor board-style prognostic deliberation.

\subsubsection{\our-Longitudinal}

\paragraph{Dataset} We curated a subset of 40 patient cases from the \msk dataset (CC BY-NC-ND 4.0) \citep{msk}, a clinicogenomic resource of cancer patients linking genomic profiles with structured clinical timelines, each with an average of five associated files, including copy-number alterations, somatic mutations, specimen pathology reports, and clinical timelines. The timelines capture key clinical events such as diagnostic procedures and treatment transitions, and are segmented into decision-relevant timepoints to support temporally grounded evaluation.

\paragraph{Q\&A design} We manually construct 183 question-answer pairs (\ie, on average 4.6 questions per patient), with clinical feedback, targeting outcome prediction, recurrence risk, and treatment progression across clinically actionable stages.

\paragraph{Tasks}
\looseness -1 The longitudinal track challenges agents to reason over temporally structured patient data segmented into decision-relevant timepoints. Initial questions assess diagnosis and disease trajectory, followed by outcome prediction (e.g., survival), recurrence forecasting, and treatment progression mapping. Genomic data—such as somatic mutations and copy---number alterations—are introduced at key stages, enabling reasoning about resistance patterns or post-treatment stratification. Agents must align treatment regimens with outcomes and integrate evolving context (e.g., updated timelines, new genomic tests, surgical pathology) to justify predictions. This setup mirrors the longitudinal deliberations of MTBs, where clinicians revise hypotheses in light of new events and cumulative history.

\subsection{Agent System}
Current LLM-based systems struggle to reason across multiple modalities and timepoints \citep{Hager2024.01.26.24301810, AlSaad_Abd-Alrazaq_Boughorbel_Ahmed_Renault_Damseh_Sheikh_2024}—a critical requirement in real-world clinical decision-making. In particular, tasks encountered in molecular tumor boards involve dynamic access to evolving patient data, integration of heterogeneous sources such as pathology, lab tests, and genomics, and the ability to contextualize findings over time. Static, single-shot prompting falls short in such settings.

To overcome these limitations, \our implements an agentic framework that enables interactive, multi-turn decision-making. Agents must actively select which files to access, manage evolving memory across turns. A key novelty of our setup is the integration of domain-specific foundation models (FMs) as callable tools besides structured biomedical resources used as tool (e.g., PubMed, DrugBank). These models—trained on large corpora of pathology slides, IHC images, or clinical texts—offer rich, pretrained representations that complement the LLM's general capabilities. Rather than evaluating FMs in isolation, \our enables agents to selectively invoke them as part of a decision-making process, simulating how clinicians consult expert resources.
This design of an agentic framework reflects how expert clinicians reason iteratively and selectively, and allows us to benchmark not only factual accuracy but also the agent’s ability to gather and use evidence in a realistic clinical workflow.

\paragraph{Agentic workflow} 
In \our, the agent engages in a multi-turn decision-making process over a temporally evolving patient trajectory. At each turn \( t \), the agent receives a clinical query \( q_t \) along with access to a set of modality-specific files \( \mathcal{F}_t = \{f_t^1, f_t^2, \ldots, f_t^k\} \), which may include digital pathology images, lab results, clinical notes, or structured genomic and temporal data. The agent may issue a request \( \mathcal{R}_t \subseteq \mathcal{F}_t \) to retrieve any subset of these files, which remain accessible only within the current turn. Namely, they do not persist across turns. However, any file from \( \mathcal{F}_t \) may be re-requested at a future turn \( t' > t \), simulating realistic constraints in clinical workflows where information must be actively re-accessed. The agent's internal memory consists of its reasoning history \( h_t \) and a record of previously accessed files \( \mathcal{R}_{\leq t} \), forming the basis for answering downstream queries. In the longitudinal track, clinical context is further enriched by an evolving timeline \( \mathcal{T}_t = \bigcup_{i=1}^{t} \tau_i \), incrementally summarizing patient history. This setup enforces non-persistent access patterns while encouraging deliberate information gathering and reasoning over temporally non-stationary data. An extensive overview of this workflow is provided in Appendix~\ref{sec:agentic_workflow}.

\paragraph{Overview of models} We select a wide range of models with varying sizes. For the multimodal part of our benchmark, we evaluate the vision-text models (including some models with reasoning capabilities): \gemma{12}, \gemma{27}, \gpt, o4-mini (reasoning), \internvl{38}, \internvl{78}, \llamavision{90}, \mistralsmall{24}, \qwenvl{7}, and \qwenvl{32}. For the longitudinal part, we evaluate a mix of text-only and vision-text models: \gemma{12}, \gemma{27}, \gpt{}, \ofourmini{} (reasoning), \llamaold{8}, \llama{70}, \qwennew{8} (reasoning), and \qwennew{32} (reasoning).

\begin{table}[t]
\centering
\caption{\footnotesize Mean accuracy and 95\% confidence intervals of various LLMs by task, estimated via bootstrap resampling. Each cell reports the model's mean accuracy, with confidence intervals computed by resampling (with replacement) 1,000 times from the set of patient–question pairs within each task.}
\resizebox{\textwidth}{!}{%
\begin{tabular}{lcccc}
\toprule
\textbf{Multi-Modal Analysis} & \textbf{Digital Pathology} & \textbf{Hematology} & \textbf{Outcome} and \textbf{Recurrence} & \textbf{Overall} \\
\midrule
\gemma{12} & 55.9 $\pm$ \scriptsize{6.4} & 74.9 $\pm$ \scriptsize{8.7} & 53.6 $\pm$ \scriptsize{13.5} & 61.5 $\pm$ \scriptsize{10.1} \\
\gemma{27} & 51.8 $\pm$ \scriptsize{6.4} & 76.9 $\pm$ \scriptsize{8.2} & 42.1 $\pm$ \scriptsize{13.5} & 56.9 $\pm$ \scriptsize{16.5} \\
\gpt & 63.2 $\pm$ \scriptsize{6.0} & 76.9 $\pm$ \scriptsize{7.7} & 59.9 $\pm$ \scriptsize{13.5} & 66.7 $\pm$ \scriptsize{8.1} \\
\ofourmini & 59.5 $\pm$ \scriptsize{6.4} & 77.8 $\pm$ \scriptsize{8.2} & 55.7 $\pm$ \scriptsize{14.4} & 64.3 $\pm$ \scriptsize{10.5} \\
\internvl{38} & 54.7 $\pm$ \scriptsize{6.4} & 79.7 $\pm$ \scriptsize{8.2} & 55.9 $\pm$ \scriptsize{13.5} & 63.5 $\pm$ \scriptsize{11.9} \\
\internvl{78} & 62.0 $\pm$ \scriptsize{6.4} & 79.7 $\pm$ \scriptsize{7.7} & 65.6 $\pm$ \scriptsize{11.5} & 69.1 $\pm$ \scriptsize{8.4} \\
\llamavision{90} & 54.6 $\pm$ \scriptsize{6.2} & 82.8 $\pm$ \scriptsize{7.2} & 51.7 $\pm$ \scriptsize{13.5} & 63.0 $\pm$ \scriptsize{14.8} \\
\mistralsmall{24} & 62.4 $\pm$ \scriptsize{6.2} & 75.8 $\pm$ \scriptsize{8.7} & 51.7 $\pm$ \scriptsize{13.5} & 63.3 $\pm$ \scriptsize{11.5} \\
\qwenvl{7} & 42.3 $\pm$ \scriptsize{6.2} & 61.1 $\pm$ \scriptsize{9.1} & 53.9 $\pm$ \scriptsize{12.5} & 52.4 $\pm$ \scriptsize{9.0} \\
\qwenvl{32} & 53.3 $\pm$ \scriptsize{6.2} & 73.0 $\pm$ \scriptsize{8.7} & 63.6 $\pm$ \scriptsize{12.5} & 63.3 $\pm$ \scriptsize{9.3} \\
\midrule
\midrule
\textbf{Longitudinal Analysis} & \textbf{Outcome} & \textbf{Progression} & \textbf{Recurrence} & \textbf{Overall} \\
\midrule
\gemma{12} & 63.3 $\pm$ \scriptsize{11.3} & 55.9 $\pm$ \scriptsize{13.2} & 54.6 $\pm$ \scriptsize{12.8} & 58.0 $\pm$ \scriptsize{4.1} \\
\gemma{27} & 57.7 $\pm$ \scriptsize{11.3} & 50.7 $\pm$ \scriptsize{14.0} & 47.4 $\pm$ \scriptsize{13.6} & 51.9 $\pm$ \scriptsize{4.9} \\
\gpt & 72.9 $\pm$ \scriptsize{10.6} & 64.8 $\pm$ \scriptsize{13.2} & 54.8 $\pm$ \scriptsize{13.6} & 64.2 $\pm$ \scriptsize{8.6} \\
\ofourmini & 66.0 $\pm$ \scriptsize{10.6} & 63.1 $\pm$ \scriptsize{12.3} & 51.1 $\pm$ \scriptsize{13.6} & 60.0 $\pm$ \scriptsize{7.1} \\
\llamaold{8} & 60.4 $\pm$ \scriptsize{11.3} & 49.0 $\pm$ \scriptsize{12.3} & 45.5 $\pm$ \scriptsize{13.6} & 51.6 $\pm$ \scriptsize{7.1} \\
\llama{70} & 73.2 $\pm$ \scriptsize{9.9} & 68.2 $\pm$ \scriptsize{13.2} & 56.7 $\pm$ \scriptsize{13.6} & 66.0 $\pm$ \scriptsize{7.8} \\
\qwennew{8} & 63.1 $\pm$ \scriptsize{11.3} & 57.6 $\pm$ \scriptsize{13.2} & 47.4 $\pm$ \scriptsize{12.7} & 56.0 $\pm$ \scriptsize{7.5} \\
\qwennew{32} & 83.0 $\pm$ \scriptsize{9.2} & 63.3 $\pm$ \scriptsize{12.3} & 54.6 $\pm$ \scriptsize{13.6} & 67.0 $\pm$ \scriptsize{13.5} \\
\bottomrule
\vspace{-10pt}

\end{tabular}
}
\label{tab:main_benchmarks}
\end{table}                                

\subsubsection{Foundation Model-based Tools}

While large language models excel at reasoning over textual inputs, they exhibit well-known limitations in visual understanding—especially when interpreting high-resolution biomedical imagery such as histopathology slides \citep{lu_multimodal_2024}. In clinical contexts like MTBs, however, the ability to analyze and contextualize pathology images is essential. At the same time, recent advances in vision-language foundation models have shown that pretrained models trained on large-scale medical imaging corpora can capture powerful, domain-specific visual representations \citep{vaidya_he_2025}.
To harness these capabilities, \our integrates foundation models as external tools (taking inspiration from \citep{schick2023toolformerlanguagemodelsteach, yao2023reactsynergizingreasoningacting}) that LLM agents can call on-demand. These models are not used passively; instead, agents learn to query them selectively as part of a broader reasoning process. This setup reflects real-world clinical workflows, where specialists consult diagnostic systems or reference image atlases to refine decisions. By exposing FMs as callable components, \our enables systematic evaluation of how agents can leverage visual expertise to complement their textual reasoning.

\subsubsection{Digital Pathology Foundation Models}

For \textbf{H\&E images}, we integrate \textsc{CONCH} \citep{lu2024avisionlanguage}, a vision-language model pretrained on over 1.17 million H\&E image–caption pairs. \textsc{CONCH} generates dual visual and textual embeddings, allowing image–text similarity computations. We expose this capability to the LLM by framing it as a tool: given an image and a list of candidate textual descriptors, the tool returns the one with highest embedding similarity to the image, based on dot product in the shared representation space.

For \textbf{IHC images}, we develop a custom tool that combines foundation model embeddings with weakly supervised learning to support quantification of marker-specific staining. Tissue regions are segmented and tiled into fixed-size patches ($256 \times 256$), each embedded using the \textsc{UNI2} foundation model \citep{chen2024uni} to produce 1536-dimensional representations. These embeddings are aggregated using an attention-based multiple instance learning (ABMIL) \citep{pmlr-v80-ilse18a} model trained to regress the percentage of positively stained cells. The ABMIL model is trained on a manually curated dataset of IHC images annotated via QuPath (see Appendix~\ref{sec:abmil}), providing marker-level supervision without requiring single-cell labels. 

\subsubsection{Analysis and Knowledge Database Tools}
To support reasoning over temporal sequences of clinical events, we introduce two tools that provide external biomedical knowledge for answering longitudinal questions more accurately: a literature search tool and a pharmacological knowledge base.

\paragraph{PubMed} The tool enables the LLM to issue natural language queries to retrieve biomedical literature relevant to a patient’s clinical trajectory. The LLM issues natural language queries, which are used to retrieve the top 30 PubMed articles. These are reranked using the \texttt{BAAI-bge-reranker-v2-m3 model} \citep{li2023making}, and the top 3 abstracts are returned to the LLM, supporting evidence-grounded reasoning for questions involving treatment effectiveness, sequencing, or disease progression.

\paragraph{DrugBank} To augment drug-related knowledge, we integrate information from DrugBank \citep{wishart_drugbank_2006, drugbank2}. When processing a patient’s clinical timeline, drug mentions are automatically linked to corresponding DrugBank entries. Relevant metadata such as therapeutic indications, mechanisms of action, and known drug interactions is incorporated into the model’s context. This enrichment enables the language model to reason about treatment sequences with greater specificity, especially in scenarios involving therapeutic decision making and longitudinal disease management.

\begin{figure}[t]
    \centering
    \includegraphics[width=0.99\linewidth]{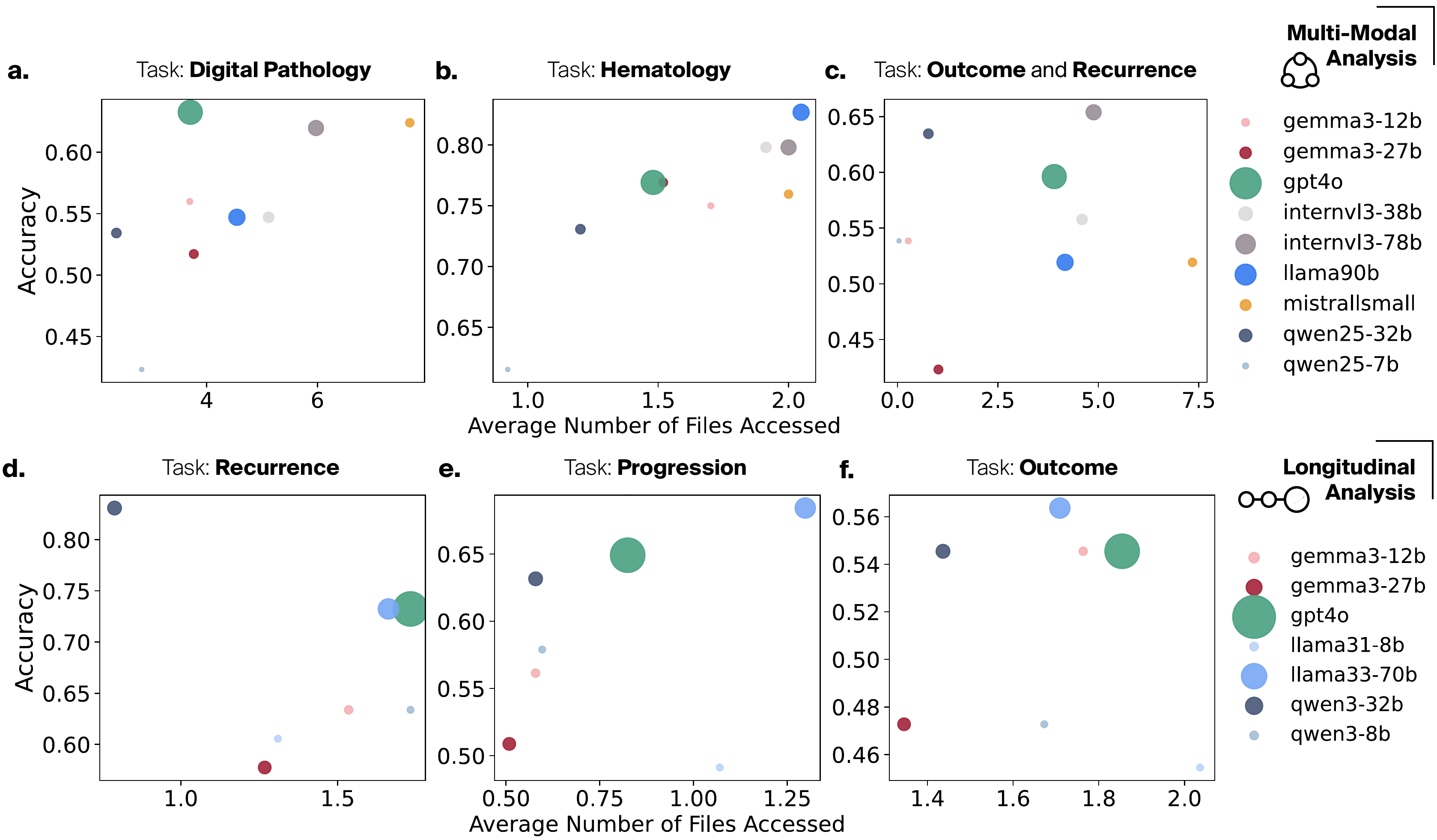}
    \caption{\footnotesize \textbf{Accuracy vs. average number of files accessed per question.} Analyzed across tasks for multi-modal understanding (\textbf{a}–\textbf{c}) and longitudinal reasoning (\textbf{d}–\textbf{f}). Each point represents a model evaluated on a specific task across all patients. Dots indicate model sizes (gpt-4o's size has been reduced for visibility). Higher file access generally correlates with increased accuracy, highlighting the importance of cross-modality and temporal integration for performance. \vspace{-10pt}}
    \label{fig:combined_dots}
\end{figure}

\vspace{-5pt}
\section{Empirical Evaluation} \label{sec:evaluation}
\vspace{-5pt}

We evaluate models on both the multi-modal and longitudinal tracks of \our under two conditions: (i) baseline inference without tool support and (ii) augmented inference with access to domain-specific tools, across several metrics, including their accuracy, analysis on multi-modal understanding and ability to reason across temporally-resolved data. Each model acts as an agent interacting with the benchmark via multi-turn dialogues, selectively retrieving and reasoning over available patient files to answer clinical questions. Models are provided only with patient metadata and a list of modality-specific files at each turn. They must request specific files and construct their answers from retrieved content. 

\subsection{Results on \our without Tools} \label{sec:results_no_tools}

\paragraph{Settings}
We evaluate a diverse set of LLMs and VLMs across all tasks in both the multimodal and longitudinal tracks of \our. To simulate realistic tumor board conditions, we adjust the context and available files for each question type, ensuring that only data plausibly accessible at the corresponding clinical stage is provided. No tools or external resources are available in this setting. Models receive only demographic details, pathology reports, imaging references, and structured clinical information. The core task remains multiple-choice question answering, but success requires multimodal reasoning, data retrieval, and longitudinal inference rather than simple pattern recognition.

\paragraph{Evaluation metrics} \label{sec:eval_metrics}
We report mean accuracy per model and task. To quantify uncertainty, we estimate 95\% confidence intervals using bootstrap resampling with 1000 iterations. For each task, we sample with replacement from the set of question outcomes per model, compute the mean accuracy per sample, and extract the 2.5th and 97.5th percentiles to define the confidence interval.
To assess their ability to incorporate findings from several modalities, we analyze the number of modalities queried compared to the resulting achieved accuracy. The analysis is conducted across 26 patients for \our-Multimodal and 40 for \our-Longitudinal. For details, see Appendix~\ref{sec:exp_details}.

\paragraph{Results for \our-Multimodal}
Accuracy across all models (of different parameter sizes) are displayed in Table~\ref{tab:main_benchmarks}. Model performance varies substantially across the multimodal tasks. Digital pathology, despite being visually complex, does not show a consistent benefit from model size—for instance, \gemma{12} outperforms its larger counterpart \gemma{27}. Hematology emerges as the most approachable task, likely due to its structured and interpretable inputs. In contrast, outcome and recurrence prediction remain the most difficult, with accuracies near random (50\%), even for leading models. The strongest overall performance is achieved by \internvl{78} at 69.1\% accuracy, outperforming closed-source baselines like \gpt\ by 2.5\%. Nevertheless, large performance gaps persist: up to 36.7\% in digital pathology, 17.2\% in hematology, and 34.6\% in outcome and recurrence prediction—highlighting the need for more robust multimodal reasoning.

\looseness-1 Instead of model size, a stronger signal emerges in the relationship between performance and the number of files accessed (Fig.~\ref{fig:combined_dots}a-c), in both \our-Multimodal and \our-Longitudinal. This suggests that effective information gathering, rather than raw scale, is a key determinant of accuracy. 
This is also demonstrated in Example~\ref{fig:example_1}: Compared to \qwenvl{7}, \gpt\ accesses more modalities including higher resolution H\&E regions-of-interest, resulting in the correct cancer subtype identification. In Example~\ref{fig:example_2}, \mistralsmall{24} requests file access to more IHCs and the H\&E slide compared to \gemma{27}, resulting in correct cancer subtype identification. 
However, this trend does not hold for outcome and recurrence tasks, where high error rates persist across models. We hypothesize that these questions require contextual grounding and biomarker interpretation beyond the current capabilities of uni-modal or tool-free agents. For further results, see Appendix~\ref{sec:further_results} and Figs.~\ref{fig:violins}).

\paragraph{Results for \our-Longitudinal} 
The longitudinal track reveals persistent weaknesses in baseline LLMs. While outcome prediction shows some promise—\qwennew{32} reaches 83.1\% accuracy—recurrence and progression tasks remain near chance (Table~\ref{tab:main_benchmarks}). Similarly, querying multiple modalities improves model performance (Fig.~\ref{fig:combined_dots}d-f). This is also demonstrated in Example~\ref{fig:example_3}: compared to \gemma{27}, \qwennew{32} re-accesses part of the patient timeline of events as well as pathological data, resulting in better cancer progression prediction. 
This disparity suggests models can detect coarse survival signals but struggle with more nuanced temporal reasoning, reflecting varied evidence complexity across tasks.
\begin{figure}
    \centering
    \includegraphics[width=0.99\linewidth]{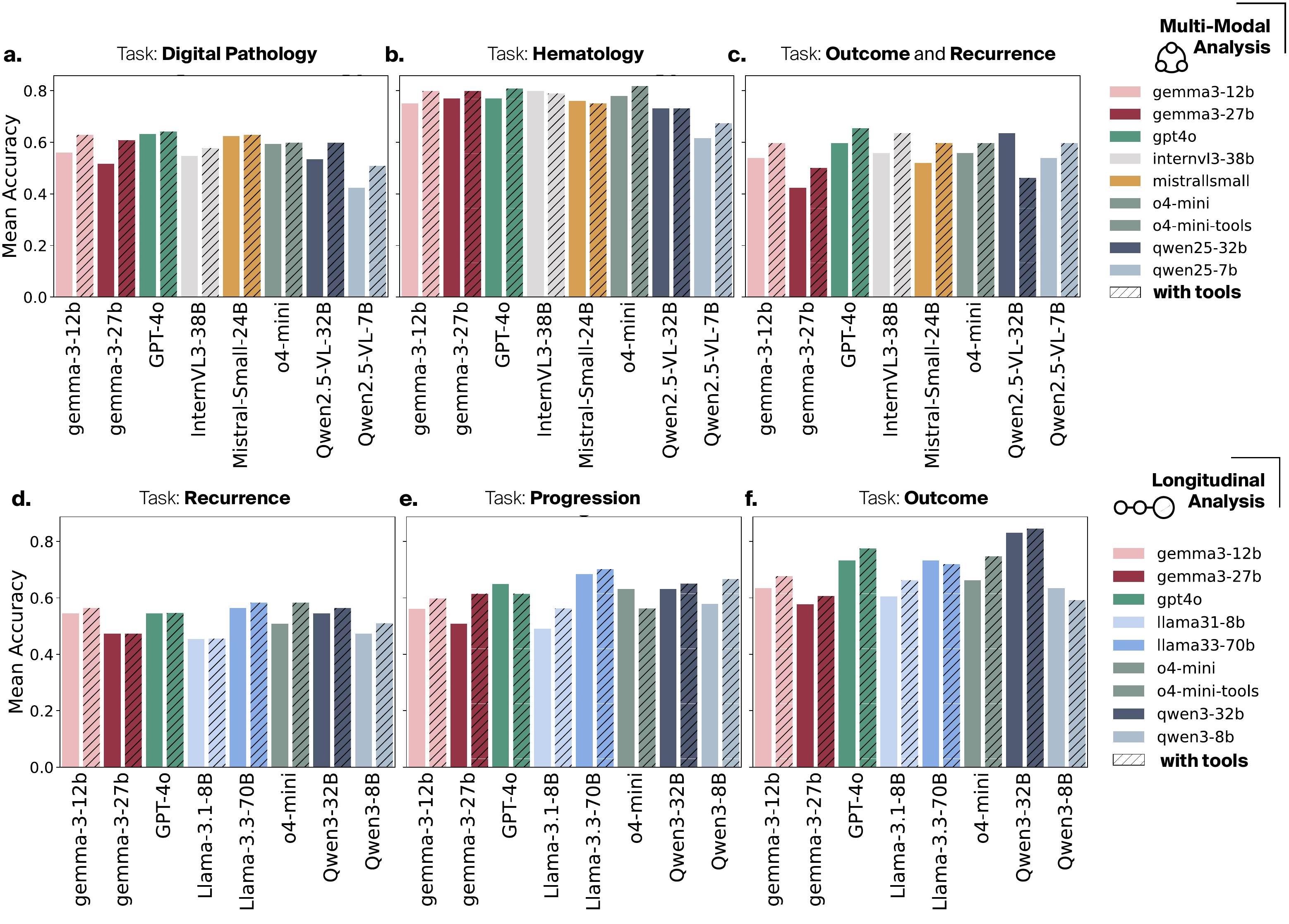}
    \caption{\footnotesize \textbf{Accuracy across models and tasks for naive and tool-augmented agents.} For multi-modal (\textbf{a.}–\textbf{c.}) and longitudinal (\textbf{d.}–\textbf{f.}) evaluation. Models equipped with tool access (hatched bars) generally show improved accuracy, highlighting the benefit of querying external resources in both multi-modal and temporal settings. \vspace{-20pt}}
    \label{fig:combined_tools}
\end{figure}

\subsection{Results on \our with Tools}

\paragraph{Settings} 
In this setup, we augment the baseline LLMs with access to external tools, including foundation model-based modules (UNI, \citet{chen2024uni} and CONCH, \citet{CONCH}) and classical biomedical resources (PubMed and DrugBank). These tools are accessible via API-style interfaces, allowing agents to retrieve structured outputs during multi-turn reasoning. Visual tools are primarily used in multimodal tasks, while longitudinal tools support reasoning over treatment history, drug interactions, and literature-based evidence.

\paragraph{Evaluation metrics} 
We use the same accuracy metrics described in Section~\ref{sec:results_no_tools}. Results are reported per task and model, and improvements are measured relative to tool-free baselines.

\paragraph{Results for \our-Multimodal}
As shown in Figure~\ref{fig:combined_tools}a–c (exact numbers in Fig.~\ref{fig:delta_bootstrap_mean}), access to visual foundation model tools significantly improves performance on all multimodal tasks. Digital pathology, in particular, benefits from integration of the FMs, with models like \gemma{12} and \qwenvl{7} showing improvements of up to 9\%. Notably, tool augmentation also improves performance on hematology tasks, despite the absence of dedicated tools for lab analysis. We attribute this to better contextual grounding: agents are more effective at integrating diverse file types when given richer information from related modalities. Outcome and recurrence tasks, which showed low baseline performance, also benefit modestly from the improved visual reasoning capabilities.

This behavior is also demonstrated through an example: in Example~\ref{fig:example_4}, \mistralsmall{24} \withtools\ in contrast to \mistralsmall{24} without tools accesses a digital pathology FM for IHC analysis. Despite both models accessing the same number of modalities, \mistralsmall{24} \withtools\ is able to properly address the question, i.e., which marker is critical in a given pathological analysis. 

\paragraph{Results for \our-Longitudinal} 
Figures~\ref{fig:combined_tools}d–f show that tool access improves performance across most longitudinal tasks, though gains are generally modest. This is expected, as no specialized foundation model currently exists for longitudinal clinical reasoning---a key limitation in this track. Instead, agents rely on general-purpose tools like DrugBank and PubMed, which still provide some benefit by enriching context and supporting evidence-based decisions. For example, progression and recurrence predictions improve by over 5\% in selected models. Outcome prediction---which already exhibited strong baseline performance—also sees incremental gains, underscoring the potential of even generic tools to enhance structured clinical reasoning.

\section{Conclusion} \label{sec:conclusion}

\our introduces a benchmark and agentic framework for evaluating AI agents in longitudinal, multi-modal oncology workflows, modeled on the structure of real molecular tumor boards. By combining temporally evolving patient data, expert-validated clinical questions, and access to external tools and pretrained foundation models, \our enables a rigorous assessment of agents' ability to reason across modalities and time.
Our evaluation shows that agents perform significantly better when equipped to query diverse modalities and leverage domain-specific models, underscoring the need for flexible, tool-augmented reasoning in clinical AI. \our shifts the field from static, uni-modal evaluation toward dynamic, decision-centric assessment grounded in clinical complexity.
While \our simulates realistic decision-making, it remains a controlled offline benchmark—agents are not yet tested in interactive, real-world clinical workflows or exposed to ambiguous or incomplete inputs requiring clarification or adaptive strategies.
Looking forward, future work will explore extending \our to more diverse clinical domains and incorporating interactive elements—paving the way for evaluating agents not only as reasoning tools but as potential collaborators in real-world precision oncology. A particularly promising direction we would like to explore involves the integration of medical foundation models with capabilities for analyzing complex longitudinal data, enabling deeper temporal reasoning and personalized decision support.

\paragraph{Societal impacts} \our offers clear benefits by promoting clinically relevant evaluation of AI agents, potentially advancing more trustworthy decision support in oncology. It encourages models to reason over multimodal, longitudinal data, closer to real-world needs. However, risks include misuse of the benchmark to suggest clinical readiness, and reduced transparency when agents rely heavily on tools. We emphasize that \our is for research only, not clinical deployment, and should be used with proper oversight and ethical safeguards.

\section{Acknowledgements}
The authors would like to thank Stavros Pantelakos, Martha Nifora, Jan Rüschoff, and Claudio de Vito for their valuable feedback for validation of the benchmark.

\bibliographystyle{unsrtnat}
\bibliography{bib}

\begin{thebibliography}{70}
\providecommand{\natexlab}[1]{#1}
\providecommand{\url}[1]{\texttt{#1}}
\expandafter\ifx\csname urlstyle\endcsname\relax
  \providecommand{\doi}[1]{doi: #1}\else
  \providecommand{\doi}{doi: \begingroup \urlstyle{rm}\Url}\fi

\bibitem[Moor et~al.(2023)Moor, Banerjee, Abad, Krumholz, Leskovec, Topol, and Rajpurkar]{moor2023foundation}
Michael Moor, Oishi Banerjee, Zahra Shakeri~Hossein Abad, Harlan~M Krumholz, Jure Leskovec, Eric~J Topol, and Pranav Rajpurkar.
\newblock Foundation models for generalist medical artificial intelligence.
\newblock \emph{Nature}, 616\penalty0 (7956):\penalty0 259--265, 2023.

\bibitem[Lu et~al.(2024{\natexlab{a}})Lu, Chen, Williamson, Chen, Zhao, Chow, Ikemura, Kim, Pouli, Patel, Soliman, Chen, Ding, Wang, Gerber, Liang, Le, Parwani, Weishaupt, and Mahmood]{Lu2024}
Ming~Y. Lu, Bowen Chen, Drew F.~K. Williamson, Richard~J. Chen, Melissa Zhao, Aaron~K. Chow, Kenji Ikemura, Ahrong Kim, Dimitra Pouli, Ankush Patel, Amr Soliman, Chengkuan Chen, Tong Ding, Judy~J. Wang, Georg Gerber, Ivy Liang, Long~Phi Le, Anil~V. Parwani, Luca~L. Weishaupt, and Faisal Mahmood.
\newblock A multimodal generative ai copilot for human pathology.
\newblock \emph{Nature}, 634\penalty0 (8033):\penalty0 466--473, Oct 2024{\natexlab{a}}.
\newblock ISSN 1476-4687.
\newblock \doi{10.1038/s41586-024-07618-3}.

\bibitem[Dai et~al.(2025)Dai, Zhang, Yang, Xu, Shen, Xia, and Wang]{Dai2025}
Dawei Dai, Yuanhui Zhang, Qianlan Yang, Long Xu, Xiaojing Shen, Shuyin Xia, and Guoyin Wang.
\newblock Pathologyvlm: a large vision-language model for pathology image understanding.
\newblock \emph{Artificial Intelligence Review}, 58\penalty0 (6):\penalty0 186, Mar 2025.
\newblock ISSN 1573-7462.
\newblock \doi{10.1007/s10462-025-11190-1}.

\bibitem[Lu et~al.(2024{\natexlab{b}})Lu, Chen, Williamson, Chen, Liang, Ding, Jaume, Odintsov, Le, Gerber, Parwani, Zhang, and Mahmood]{CONCH}
Ming~Y. Lu, Bowen Chen, Drew F.~K. Williamson, Richard~J. Chen, Ivy Liang, Tong Ding, Guillaume Jaume, Igor Odintsov, Long~Phi Le, Georg Gerber, Anil~V. Parwani, Andrew Zhang, and Faisal Mahmood.
\newblock A visual-language foundation model for computational pathology.
\newblock \emph{Nature Medicine}, 30\penalty0 (3):\penalty0 863--874, Mar 2024{\natexlab{b}}.
\newblock ISSN 1546-170X.
\newblock \doi{10.1038/s41591-024-02856-4}.

\bibitem[Choudhuri et~al.(2025)Choudhuri, Polgreen, Segre, and Adhikari]{Choudhuri2025.01.19.25320797}
Akash Choudhuri, Philip Polgreen, Alberto Segre, and Bijaya Adhikari.
\newblock Summarizing clinical notes using llms for icu bounceback and length-of-stay prediction.
\newblock \emph{medRxiv}, 2025.
\newblock \doi{10.1101/2025.01.19.25320797}.

\bibitem[Yang et~al.(2024)Yang, Mitra, Kwon, and Yu]{yang2024clinicalmambagenerativeclinicallanguage}
Zhichao Yang, Avijit Mitra, Sunjae Kwon, and Hong Yu.
\newblock Clinicalmamba: A generative clinical language model on longitudinal clinical notes.
\newblock \emph{arXiv preprint arXiv:2403.05795}, 2024.

\bibitem[Schmidgall et~al.(2024)Schmidgall, Ziaei, Harris, Reis, Jopling, and Moor]{schmidgall2024agentclinicmultimodalagentbenchmark}
Samuel Schmidgall, Rojin Ziaei, Carl Harris, Eduardo Reis, Jeffrey Jopling, and Michael Moor.
\newblock Agentclinic: a multimodal agent benchmark to evaluate ai in simulated clinical environments.
\newblock \emph{arXiv preprint arXiv:2405.07960}, 2024.

\bibitem[Wang et~al.(2025{\natexlab{a}})Wang, Ma, Wang, Wu, Chen, Li, and Yuan]{wang2025surveyllmbasedagentsmedicine}
Wenxuan Wang, Zizhan Ma, Zheng Wang, Chenghan Wu, Wenting Chen, Xiang Li, and Yixuan Yuan.
\newblock A survey of llm-based agents in medicine: How far are we from baymax?
\newblock \emph{arXiv preprint arXiv:2502.11211}, 2025{\natexlab{a}}.

\bibitem[Wang et~al.(2024{\natexlab{a}})Wang, Wang, Danek, Li, Mack, Poon, Wang, Rajpurkar, and Sun]{wang2024perspectiveadaptinggeneralistai}
Zifeng Wang, Hanyin Wang, Benjamin Danek, Ying Li, Christina Mack, Hoifung Poon, Yajuan Wang, Pranav Rajpurkar, and Jimeng Sun.
\newblock A perspective for adapting generalist ai to specialized medical ai applications and their challenges.
\newblock \emph{arXiv preprint arXiv:2411.00024}, 2024{\natexlab{a}}.

\bibitem[Wang et~al.(2025{\natexlab{b}})Wang, Schmidgall, Jaeger, Zhang, Pilgrim, Matias, Barral, Fleet, and Azizi]{wang2025txgemmaefficientagenticllms}
Eric Wang, Samuel Schmidgall, Paul~F. Jaeger, Fan Zhang, Rory Pilgrim, Yossi Matias, Joelle Barral, David Fleet, and Shekoofeh Azizi.
\newblock Txgemma: Efficient and agentic llms for therapeutics.
\newblock \emph{arXiv preprint arXiv:2504.06196}, 2025{\natexlab{b}}.

\bibitem[Gao et~al.(2024)Gao, Fang, Huang, Giunchiglia, Noori, Schwarz, Ektefaie, Kondic, and Zitnik]{gao2024empoweringbiomedicaldiscoveryai}
Shanghua Gao, Ada Fang, Yepeng Huang, Valentina Giunchiglia, Ayush Noori, Jonathan~Richard Schwarz, Yasha Ektefaie, Jovana Kondic, and Marinka Zitnik.
\newblock Empowering biomedical discovery with ai agents.
\newblock \emph{arXiv preprint arXiv:2404.02831}, 2024.

\bibitem[Lee et~al.(2024)Lee, Ferber, Rood, Regev, and Kather]{Lee2024}
Yongju Lee, Dyke Ferber, Jennifer~E. Rood, Aviv Regev, and Jakob~Nikolas Kather.
\newblock How ai agents will change cancer research and oncology.
\newblock \emph{Nature Cancer}, 5\penalty0 (12):\penalty0 1765--1767, Dec 2024.
\newblock ISSN 2662-1347.
\newblock \doi{10.1038/s43018-024-00861-7}.

\bibitem[Yue et~al.(2024)Yue, Xing, Chen, and Fu]{yue2024clinicalagentclinicaltrialmultiagent}
Ling Yue, Sixue Xing, Jintai Chen, and Tianfan Fu.
\newblock Clinicalagent: Clinical trial multi-agent system with large language model-based reasoning.
\newblock \emph{arXiv preprint arXiv:2404.14777}, 2024.

\bibitem[Fallahpour et~al.(2025)Fallahpour, Ma, Munim, Lyu, and Wang]{fallahpour2025medraxmedicalreasoningagent}
Adibvafa Fallahpour, Jun Ma, Alif Munim, Hongwei Lyu, and Bo~Wang.
\newblock Medrax: Medical reasoning agent for chest x-ray.
\newblock \emph{arXiv preprint arXiv:2502.02673}, 2025.

\bibitem[Tsimberidou et~al.(2023)Tsimberidou, Kahle, Vo, Baysal, Johnson, and Meric-Bernstam]{Tsimberidou2023}
Apostolia~M Tsimberidou, Michael Kahle, Henry~Hiep Vo, Mehmet~A Baysal, Amber Johnson, and Funda Meric-Bernstam.
\newblock Molecular tumour boards—current and future considerations for precision oncology.
\newblock \emph{Nature Reviews Clinical Oncology}, 20\penalty0 (12):\penalty0 843--863, 2023.

\bibitem[Jiang et~al.(2025)Jiang, Black, Geng, Park, Zou, Ng, and Chen]{jiang2025medagentbenchrealisticvirtualehr}
Yixing Jiang, Kameron~C. Black, Gloria Geng, Danny Park, James Zou, Andrew~Y. Ng, and Jonathan~H. Chen.
\newblock Medagentbench: A realistic virtual ehr environment to benchmark medical llm agents.
\newblock \emph{arXiv preprint arXiv:2501.14654}, 2025.

\bibitem[Li et~al.(2024)Li, Balachandran, Feng, Ilgen, Pierson, Koh, and Tsvetkov]{li2024mediqquestionaskingllmsbenchmark}
Shuyue~Stella Li, Vidhisha Balachandran, Shangbin Feng, Jonathan~S Ilgen, Emma Pierson, Pang~Wei Koh, and Yulia Tsvetkov.
\newblock Mediq: Question-asking llms and a benchmark for reliable interactive clinical reasoning.
\newblock In \emph{Advances in Neural Information Processing Systems (NeurIPS)}, volume~38, 2024.

\bibitem[Wu et~al.(2024)Wu, Zhao, Zhang, Wu, Zhu, Zhang, Ouyang, Zhang, Wang, Lin, Yang, Zhao, and Zheng]{medjourney}
Xian Wu, Yutian Zhao, Yunyan Zhang, Jiageng Wu, Zhihong Zhu, Yingying Zhang, Yi~Ouyang, Ziheng Zhang, Huimin Wang, Zhenxi Lin, Jie Yang, Shuang Zhao, and Yefeng Zheng.
\newblock Medjourney: Benchmark and evaluation of large language models over patient clinical journey.
\newblock In A.~Globerson, L.~Mackey, D.~Belgrave, A.~Fan, U.~Paquet, J.~Tomczak, and C.~Zhang, editors, \emph{Advances in Neural Information Processing Systems}, volume~37, pages 87621--87646. Curran Associates, Inc., 2024.

\bibitem[Kweon et~al.(2024)Kweon, Kim, Kwak, Cha, Yoon, Kim, Yang, Won, and Choi]{kweon2024ehrnoteqa}
Sunjun Kweon, Jiyoun Kim, Heeyoung Kwak, Dongchul Cha, Hangyul Yoon, Kwanghyun Kim, Jeewon Yang, Seunghyun Won, and Edward Choi.
\newblock Ehrnoteqa: An llm benchmark for real-world clinical practice using discharge summaries.
\newblock \emph{arXiv preprint arXiv:2402.16040}, 2024.

\bibitem[Chen et~al.(2023)Chen, Kansal, Chen, Jin, Reisler, Kim, and Rajpurkar]{chen_multimodal_2023}
Emma Chen, Aman Kansal, Julie Chen, Boyang~Tom Jin, Julia~Rachel Reisler, David~A. Kim, and Pranav Rajpurkar.
\newblock Multimodal {Clinical} {Benchmark} for {Emergency} {Care} ({MC}-{BEC}): {A} {Comprehensive} {Benchmark} for {Evaluating} {Foundation} {Models} in {Emergency} {Medicine}.
\newblock \emph{arXiv preprint arXiv:2311.04937}, 2023.

\bibitem[Wang et~al.(2024{\natexlab{b}})Wang, Su, Huan, Liu, Chen, Zhang, Li, Chang, Xin, Shen, et~al.]{wang2024asclepius}
Wenxuan Wang, Yihang Su, Jingyuan Huan, Jie Liu, Wenting Chen, Yudi Zhang, Cheng-Yi Li, Kao-Jung Chang, Xiaohan Xin, Linlin Shen, et~al.
\newblock Asclepius: A spectrum evaluation benchmark for medical multi-modal large language models.
\newblock \emph{arXiv preprint arXiv:2402.11217}, 2024{\natexlab{b}}.

\bibitem[Liu et~al.(2024)Liu, Li, Zhou, Yin, Yang, Tang, Luo, Zeng, Jiang, Gao, Nigam, Nag, Yin, Hua, Zhou, Rohanian, Thakur, Clifton, and Clifton]{Liu2024ClinicBench}
Fenglin Liu, Zheng Li, Hongjian Zhou, Qingyu Yin, Jingfeng Yang, Xianfeng Tang, Chen Luo, Ming Zeng, Haoming Jiang, Yifan Gao, Priyanka Nigam, Sreyashi Nag, Bing Yin, Yining Hua, Xuan Zhou, Omid Rohanian, Anshul Thakur, Lei Clifton, and David~A. Clifton.
\newblock Large language models are poor clinical decision-makers: A comprehensive benchmark.
\newblock In \emph{Conference on Empirical Methods in Natural Language Processing (EMNLP)}, 2024.

\bibitem[Jaume et~al.(2024)Jaume, Doucet, Song, Lu, Almagro-Perez, Wagner, Vaidya, Chen, Williamson, Kim, and Mahmood]{jaume2024hest}
Guillaume Jaume, Paul Doucet, Andrew~H. Song, Ming~Y. Lu, Cristina Almagro-Perez, Sophia~J. Wagner, Anurag~J. Vaidya, Richard~J. Chen, Drew F.~K. Williamson, Ahrong Kim, and Faisal Mahmood.
\newblock Hest-1k: A dataset for spatial transcriptomics and histology image analysis.
\newblock In \emph{Advances in Neural Information Processing Systems}, December 2024.

\bibitem[Mitchener et~al.(2025)Mitchener, Laurent, Tenmann, Narayanan, Wellawatte, White, Sani, and Rodriques]{mitchener2025bixbenchcomprehensivebenchmarkllmbased}
Ludovico Mitchener, Jon~M Laurent, Benjamin Tenmann, Siddharth Narayanan, Geemi~P Wellawatte, Andrew White, Lorenzo Sani, and Samuel~G Rodriques.
\newblock Bixbench: a comprehensive benchmark for llm-based agents in computational biology.
\newblock \emph{arXiv preprint arXiv:2503.00096}, 2025.

\bibitem[Ferber et~al.(2024)Ferber, Nahhas, Wölflein, Wiest, Clusmann, Leßman, Foersch, Lammert, Tschochohei, Jäger, Salto-Tellez, Schultz, Truhn, and Kather]{ferber2024autonomousartificialintelligenceagents}
Dyke Ferber, Omar S. M.~El Nahhas, Georg Wölflein, Isabella~C. Wiest, Jan Clusmann, Marie-Elisabeth Leßman, Sebastian Foersch, Jacqueline Lammert, Maximilian Tschochohei, Dirk Jäger, Manuel Salto-Tellez, Nikolaus Schultz, Daniel Truhn, and Jakob~Nikolas Kather.
\newblock Autonomous artificial intelligence agents for clinical decision making in oncology.
\newblock \emph{arXiv preprint arXiv:2404.04667}, 2024.

\bibitem[Li et~al.(2025)Li, Lai, Li, Ren, Zhang, Kang, Wang, Li, Zhang, Ma, and Liu]{li2025agenthospitalsimulacrumhospital}
Junkai Li, Yunghwei Lai, Weitao Li, Jingyi Ren, Meng Zhang, Xinhui Kang, Siyu Wang, Peng Li, Ya-Qin Zhang, Weizhi Ma, and Yang Liu.
\newblock Agent hospital: A simulacrum of hospital with evolvable medical agents.
\newblock \emph{arXiv preprint arXiv:2405.02957}, 2025.

\bibitem[D{\"o}rrich et~al.(2024)D{\"o}rrich, Balk, Heusinger, Beyer, Kanso, Matek, Hartmann, Iro, Eckstein, Gostian, and Kist]{hancock}
Marion D{\"o}rrich, Matthias Balk, Tatjana Heusinger, Sandra Beyer, Hassan Kanso, Christian Matek, Arndt Hartmann, Heinrich Iro, Markus Eckstein, Antoniu-Oreste Gostian, and Andreas~M. Kist.
\newblock A multimodal dataset for precision oncology in head and neck cancer.
\newblock \emph{medRxiv}, 2024.
\newblock \doi{10.1101/2024.05.29.24308141}.

\bibitem[Jee et~al.(2024)Jee, Fong, Pichotta, Tran, Luthra, Waters, Fu, Altoe, Liu, Maron, et~al.]{msk}
Justin Jee, Christopher Fong, Karl Pichotta, Thinh~Ngoc Tran, Anisha Luthra, Michele Waters, Chenlian Fu, Mirella Altoe, Si-Yang Liu, Steven~B Maron, et~al.
\newblock Automated real-world data integration improves cancer outcome prediction.
\newblock \emph{Nature}, pages 1--9, 2024.

\bibitem[Hager et~al.(2024)Hager, Jungmann, Bhagat, Hubrecht, Knauer, Vielhauer, Holland, Braren, Makowski, Kaisis, and Rueckert]{Hager2024.01.26.24301810}
Paul Hager, Friederike Jungmann, Kunal Bhagat, Inga Hubrecht, Manuel Knauer, Jakob Vielhauer, Robbie Holland, Rickmer Braren, Marcus Makowski, Georgios Kaisis, and Daniel Rueckert.
\newblock Evaluating and mitigating limitations of large language models in clinical decision making.
\newblock \emph{medRxiv}, 2024.
\newblock \doi{10.1101/2024.01.26.24301810}.

\bibitem[AlSaad et~al.(2024)AlSaad, Abd-Alrazaq, Boughorbel, Ahmed, Renault, Damseh, and Sheikh]{AlSaad_Abd-Alrazaq_Boughorbel_Ahmed_Renault_Damseh_Sheikh_2024}
Rawan AlSaad, Alaa Abd-Alrazaq, Sabri Boughorbel, Arfan Ahmed, Max-Antoine Renault, Rafat Damseh, and Javaid Sheikh.
\newblock Multimodal large language models in healthcare: Applications, challenges, and future outlook.
\newblock \emph{Journal of Medical Internet Research}, 26:\penalty0 e59505, August 2024.
\newblock \doi{10.2196/59505}.

\bibitem[Lu et~al.(2024{\natexlab{c}})Lu, Chen, Williamson, Chen, Zhao, Chow, Ikemura, Kim, Pouli, Patel, Soliman, Chen, Ding, Wang, Gerber, Liang, Le, Parwani, Weishaupt, and Mahmood]{lu_multimodal_2024}
Ming~Y. Lu, Bowen Chen, Drew F.~K. Williamson, Richard~J. Chen, Melissa Zhao, Aaron~K. Chow, Kenji Ikemura, Ahrong Kim, Dimitra Pouli, Ankush Patel, Amr Soliman, Chengkuan Chen, Tong Ding, Judy~J. Wang, Georg Gerber, Ivy Liang, Long~Phi Le, Anil~V. Parwani, Luca~L. Weishaupt, and Faisal Mahmood.
\newblock A multimodal generative {AI} copilot for human pathology.
\newblock \emph{Nature}, 634\penalty0 (8033):\penalty0 466--473, October 2024{\natexlab{c}}.
\newblock ISSN 1476-4687.
\newblock \doi{10.1038/s41586-024-07618-3}.
\newblock Publisher: Nature Publishing Group.

\bibitem[Vaidya et~al.(2025{\natexlab{a}})Vaidya, Zhang, Jaume, Song, Ding, Wagner, Lu, Doucet, Robertson, Almagro-Perez, Chen, ElHarouni, Ayoub, Bossi, Ligon, Gerber, Le, and Mahmood]{vaidya_he_2025}
Anurag Vaidya, Andrew Zhang, Guillaume Jaume, Andrew~H. Song, Tong Ding, Sophia~J. Wagner, Ming~Y. Lu, Paul Doucet, Harry Robertson, Cristina Almagro-Perez, Richard~J. Chen, Dina ElHarouni, Georges Ayoub, Connor Bossi, Keith~L. Ligon, Georg Gerber, Long~Phi Le, and Faisal Mahmood.
\newblock H\&{E}, {DNA}, {scRNA}-seq: {Molecular}-driven {Foundation} {Model} for {Oncologic} {Pathology}.
\newblock \emph{arXiv preprint arXiv:2501.16652}, January 2025{\natexlab{a}}.

\bibitem[Schick et~al.(2023)Schick, Dwivedi-Yu, Dessì, Raileanu, Lomeli, Zettlemoyer, Cancedda, and Scialom]{schick2023toolformerlanguagemodelsteach}
Timo Schick, Jane Dwivedi-Yu, Roberto Dessì, Roberta Raileanu, Maria Lomeli, Luke Zettlemoyer, Nicola Cancedda, and Thomas Scialom.
\newblock Toolformer: Language models can teach themselves to use tools.
\newblock \emph{arXiv preprint arXiv:2302.04761}, 2023.

\bibitem[Yao et~al.(2023)Yao, Zhao, Yu, Du, Shafran, Narasimhan, and Cao]{yao2023reactsynergizingreasoningacting}
Shunyu Yao, Jeffrey Zhao, Dian Yu, Nan Du, Izhak Shafran, Karthik Narasimhan, and Yuan Cao.
\newblock React: Synergizing reasoning and acting in language models.
\newblock \emph{arXiv preprint arXiv:2210.03629}, 2023.

\bibitem[Lu et~al.(2024{\natexlab{d}})Lu, Chen, Williamson, Chen, Liang, Ding, Jaume, Odintsov, Le, Gerber, et~al.]{lu2024avisionlanguage}
Ming~Y Lu, Bowen Chen, Drew~FK Williamson, Richard~J Chen, Ivy Liang, Tong Ding, Guillaume Jaume, Igor Odintsov, Long~Phi Le, Georg Gerber, et~al.
\newblock A visual-language foundation model for computational pathology.
\newblock \emph{Nature Medicine}, 30:\penalty0 863–874, 2024{\natexlab{d}}.

\bibitem[Chen et~al.(2024{\natexlab{a}})Chen, Ding, Lu, Williamson, Jaume, Chen, Zhang, Shao, Song, Shaban, et~al.]{chen2024uni}
Richard~J Chen, Tong Ding, Ming~Y Lu, Drew~FK Williamson, Guillaume Jaume, Bowen Chen, Andrew Zhang, Daniel Shao, Andrew~H Song, Muhammad Shaban, et~al.
\newblock Towards a general-purpose foundation model for computational pathology.
\newblock \emph{Nature Medicine}, 2024{\natexlab{a}}.

\bibitem[Ilse et~al.(2018)Ilse, Tomczak, and Welling]{pmlr-v80-ilse18a}
Maximilian Ilse, Jakub Tomczak, and Max Welling.
\newblock Attention-based deep multiple instance learning.
\newblock In Jennifer Dy and Andreas Krause, editors, \emph{Proceedings of the 35th International Conference on Machine Learning}, volume~80 of \emph{Proceedings of Machine Learning Research}, pages 2127--2136. PMLR, 10--15 Jul 2018.

\bibitem[Li et~al.(2023)Li, Liu, Xiao, and Shao]{li2023making}
Chaofan Li, Zheng Liu, Shitao Xiao, and Yingxia Shao.
\newblock Making large language models a better foundation for dense retrieval.
\newblock \emph{arXiv preprint arXiv:2312.15503}, 2023.

\bibitem[Wishart(2006)]{wishart_drugbank_2006}
D.~S. Wishart.
\newblock {DrugBank}: a comprehensive resource for in silico drug discovery and exploration.
\newblock \emph{Nucleic Acids Research}, 34\penalty0 (90001):\penalty0 D668--D672, January 2006.
\newblock ISSN 0305-1048, 1362-4962.
\newblock \doi{10.1093/nar/gkj067}.

\bibitem[Knox et~al.(2023)Knox, Wilson, Klinger, Franklin, Oler, Wilson, Pon, Cox, Chin, Strawbridge, Garcia-Patino, Kruger, Sivakumaran, Sanford, Doshi, Khetarpal, Fatokun, Doucet, Zubkowski, Rayat, Jackson, Harford, Anjum, Zakir, Wang, et~al.]{drugbank2}
Craig Knox, Mike Wilson, Christen~M Klinger, Mark Franklin, Eponine Oler, Alex Wilson, Allison Pon, Jordan Cox, Na~Eun Chin, Seth~A Strawbridge, Marysol Garcia-Patino, Ray Kruger, Aadhavya Sivakumaran, Selena Sanford, Rahil Doshi, Nitya Khetarpal, Omolola Fatokun, Daphnee Doucet, Ashley Zubkowski, Dorsa~Yahya Rayat, Hayley Jackson, Karxena Harford, Afia Anjum, Mahi Zakir, Fei Wang, et~al.
\newblock Drugbank 6.0: the drugbank knowledgebase for 2024.
\newblock \emph{Nucleic Acids Research}, 52\penalty0 (D1):\penalty0 D1265–D1275, November 2023.
\newblock \doi{10.1093/nar/gkad976}.

\bibitem[Jin et~al.(2020)Jin, Pan, Oufattole, Weng, Fang, and Szolovits]{medqa}
Di~Jin, Eileen Pan, Nassim Oufattole, Wei-Hung Weng, Hanyi Fang, and Peter Szolovits.
\newblock What disease does this patient have? a large-scale open domain question answering dataset from medical exams.
\newblock \emph{arXiv preprint arXiv:2009.13081}, 2020.

\bibitem[Chen et~al.(2024{\natexlab{b}})Chen, Ye, Wang, Li, Deng, Li, Li, Duan, Huang, Su, Wang, Zhang, Fu, Cai, Zhuang, Seibel, He, and Qiao]{chen2024gmaimmbenchcomprehensivemultimodalevaluation}
Pengcheng Chen, Jin Ye, Guoan Wang, Yanjun Li, Zhongying Deng, Wei Li, Tianbin Li, Haodong Duan, Ziyan Huang, Yanzhou Su, Benyou Wang, Shaoting Zhang, Bin Fu, Jianfei Cai, Bohan Zhuang, Eric~J Seibel, Junjun He, and Yu~Qiao.
\newblock Gmai-mmbench: A comprehensive multimodal evaluation benchmark towards general medical ai.
\newblock \emph{arXiv preprint arXiv:2408.03361}, 2024{\natexlab{b}}.

\bibitem[Xie et~al.(2024)Xie, Zhou, Gao, Wu, Li, Zhou, Liu, Xing, Zou, Xie, and Zhou]{xie2024medtrinity25mlargescalemultimodaldataset}
Yunfei Xie, Ce~Zhou, Lang Gao, Juncheng Wu, Xianhang Li, Hong-Yu Zhou, Sheng Liu, Lei Xing, James Zou, Cihang Xie, and Yuyin Zhou.
\newblock Medtrinity-25m: A large-scale multimodal dataset with multigranular annotations for medicine.
\newblock \emph{arXiv preprint arXiv:2408.02900}, 2024.

\bibitem[Bulten et~al.(2022)Bulten, Kartasalo, Chen, Ström, Pinckaers, Nagpal, Cai, Steiner, Van~Boven, Vink, De~Kaa, Van Der~Laak, Amin, Evans, Van Der~Kwast, Allan, Humphrey, Grönberg, et~al.]{panda}
Wouter Bulten, Kimmo Kartasalo, Po-Hsuan~Cameron Chen, Peter Ström, Hans Pinckaers, Kunal Nagpal, Yuannan Cai, David~F. Steiner, Hester Van~Boven, Robert Vink, Christina Hulsbergen-Van De~Kaa, Jeroen Van Der~Laak, Mahul~B. Amin, Andrew~J. Evans, Theodorus Van Der~Kwast, Robert Allan, Peter~A. Humphrey, Henrik Grönberg, et~al.
\newblock Artificial intelligence for diagnosis and gleason grading of prostate cancer: the panda challenge.
\newblock \emph{Nature Medicine}, 28\penalty0 (1):\penalty0 154–163, January 2022.
\newblock \doi{10.1038/s41591-021-01620-2}.

\bibitem[Wang et~al.(2025{\natexlab{c}})Wang, Wu, Cai, Low, Yang, Li, and Jin]{wang2025medagentproevidencebasedmultimodalmedical}
Ziyue Wang, Junde Wu, Linghan Cai, Chang~Han Low, Xihong Yang, Qiaxuan Li, and Yueming Jin.
\newblock Medagent-pro: Towards evidence-based multi-modal medical diagnosis via reasoning agentic workflow.
\newblock \emph{arXiv preprint arXiv:2503.18968}, 2025{\natexlab{c}}.

\bibitem[Zheng et~al.(2025)Zheng, Wu, Qiu, Dai, Zhang, Wang, and Xie]{zheng2025modernllmsactagent}
Qiaoyu Zheng, Chaoyi Wu, Pengcheng Qiu, Lisong Dai, Ya~Zhang, Yanfeng Wang, and Weidi Xie.
\newblock How well can modern llms act as agent cores in radiology environments?
\newblock \emph{arXiv preprint arXiv:2412.09529}, 2025.

\bibitem[Ghezloo et~al.(2025)Ghezloo, Seyfioglu, Soraki, Ikezogwo, Li, Vivekanandan, Elmore, Krishna, and Shapiro]{ghezloo2025pathfinder}
Fatemeh Ghezloo, Mehmet~Saygin Seyfioglu, Rustin Soraki, Wisdom~O Ikezogwo, Beibin Li, Tejoram Vivekanandan, Joann~G Elmore, Ranjay Krishna, and Linda Shapiro.
\newblock Pathfinder: A multi-modal multi-agent system for medical diagnostic decision-making applied to histopathology.
\newblock \emph{arXiv preprint arXiv:2502.08916}, 2025.

\bibitem[Wang et~al.(2025{\natexlab{d}})Wang, Wu, Low, and Jin]{wang2025medagent}
Ziyue Wang, Junde Wu, Chang~Han Low, and Yueming Jin.
\newblock Medagent-pro: Towards evidence-based multi-modal medical diagnosis via reasoning agentic workflow.
\newblock 2025{\natexlab{d}}.

\bibitem[Luo et~al.(2022)Luo, Sun, Xia, Qin, Zhang, Poon, and Liu]{10.1093/bib/bbac409}
Renqian Luo, Liai Sun, Yingce Xia, Tao Qin, Sheng Zhang, Hoifung Poon, and Tie-Yan Liu.
\newblock Biogpt: generative pre-trained transformer for biomedical text generation and mining.
\newblock \emph{Briefings in Bioinformatics}, 23\penalty0 (6):\penalty0 bbac409, 09 2022.
\newblock ISSN 1477-4054.
\newblock \doi{10.1093/bib/bbac409}.
\newblock URL \url{https://doi.org/10.1093/bib/bbac409}.

\bibitem[Bolton et~al.(2024)Bolton, Venigalla, Yasunaga, Hall, Xiong, Lee, Daneshjou, Frankle, Liang, Carbin, and Manning]{bolton2024biomedlm27bparameterlanguage}
Elliot Bolton, Abhinav Venigalla, Michihiro Yasunaga, David Hall, Betty Xiong, Tony Lee, Roxana Daneshjou, Jonathan Frankle, Percy Liang, Michael Carbin, and Christopher~D. Manning.
\newblock Biomedlm: A 2.7b parameter language model trained on biomedical text, 2024.
\newblock URL \url{https://arxiv.org/abs/2403.18421}.

\bibitem[Saillard et~al.(2024)Saillard, Jenatton, Llinares-López, Mariet, Cahané, Durand, and Vert]{hoptimus0}
Charlie Saillard, Rodolphe Jenatton, Felipe Llinares-López, Zelda Mariet, David Cahané, Eric Durand, and Jean-Philippe Vert.
\newblock H-optimus-0, 2024.
\newblock URL \url{https://github.com/bioptimus/releases/tree/main/models/h-optimus/v0}.

\bibitem[Filiot et~al.(2024)Filiot, Jacob, Kain, and Saillard]{filiot2024phikonv2largepublicfeature}
Alexandre Filiot, Paul Jacob, Alice~Mac Kain, and Charlie Saillard.
\newblock Phikon-v2, a large and public feature extractor for biomarker prediction, 2024.
\newblock URL \url{https://arxiv.org/abs/2409.09173}.

\bibitem[Vorontsov et~al.(2024)Vorontsov, Bozkurt, Casson, Shaikovski, Zelechowski, Severson, Zimmermann, Hall, Tenenholtz, Fusi, Yang, Mathieu, van Eck, Lee, Viret, Robert, Wang, Kunz, Lee, Bernhard, Godrich, Oakley, Millar, Hanna, Wen, Retamero, Moye, Yousfi, Kanan, Klimstra, Rothrock, Liu, and Fuchs]{Vorontsov2024}
Eugene Vorontsov, Alican Bozkurt, Adam Casson, George Shaikovski, Michal Zelechowski, Kristen Severson, Eric Zimmermann, James Hall, Neil Tenenholtz, Nicolo Fusi, Ellen Yang, Philippe Mathieu, Alexander van Eck, Donghun Lee, Julian Viret, Eric Robert, Yi~Kan Wang, Jeremy~D. Kunz, Matthew C.~H. Lee, Jan~H. Bernhard, Ran~A. Godrich, Gerard Oakley, Ewan Millar, Matthew Hanna, Hannah Wen, Juan~A. Retamero, William~A. Moye, Razik Yousfi, Christopher Kanan, David~S. Klimstra, Brandon Rothrock, Siqi Liu, and Thomas~J. Fuchs.
\newblock A foundation model for clinical-grade computational pathology and rare cancers detection.
\newblock \emph{Nature Medicine}, 30\penalty0 (10):\penalty0 2924--2935, Oct 2024.
\newblock ISSN 1546-170X.
\newblock \doi{10.1038/s41591-024-03141-0}.
\newblock URL \url{https://doi.org/10.1038/s41591-024-03141-0}.

\bibitem[Yan et~al.(2025)Yan, Wu, Li, Wang, Lu, Chen, Gao, Li, Yan, Ma, Chen, Lu, Chen, Wang, Ling, Wang, Wang, Huang, Hua, Liu, Ma, Shen, Zhang, He, Chen, Zhang, and Wang]{yan2025pathorchestracomprehensivefoundationmodel}
Fang Yan, Jianfeng Wu, Jiawen Li, Wei Wang, Jiaxuan Lu, Wen Chen, Zizhao Gao, Jianan Li, Hong Yan, Jiabo Ma, Minda Chen, Yang Lu, Qing Chen, Yizhi Wang, Xitong Ling, Xuenian Wang, Zihan Wang, Qiang Huang, Shengyi Hua, Mianxin Liu, Lei Ma, Tian Shen, Xiaofan Zhang, Yonghong He, Hao Chen, Shaoting Zhang, and Zhe Wang.
\newblock Pathorchestra: A comprehensive foundation model for computational pathology with over 100 diverse clinical-grade tasks, 2025.
\newblock URL \url{https://arxiv.org/abs/2503.24345}.

\bibitem[Xiang et~al.(2025)Xiang, Wang, Zhang, Xi, Eweje, Chen, Li, Bergstrom, Gopaulchan, Kim, Yu, Willens, Olguin, Nirschl, Neal, Diehn, Yang, and Li]{Xiang2025}
Jinxi Xiang, Xiyue Wang, Xiaoming Zhang, Yinghua Xi, Feyisope Eweje, Yijiang Chen, Yuchen Li, Colin Bergstrom, Matthew Gopaulchan, Ted Kim, Kun-Hsing Yu, Sierra Willens, Francesca~Maria Olguin, Jeffrey~J. Nirschl, Joel Neal, Maximilian Diehn, Sen Yang, and Ruijiang Li.
\newblock A vision--language foundation model for precision oncology.
\newblock \emph{Nature}, 638\penalty0 (8051):\penalty0 769--778, Feb 2025.
\newblock ISSN 1476-4687.
\newblock \doi{10.1038/s41586-024-08378-w}.
\newblock URL \url{https://doi.org/10.1038/s41586-024-08378-w}.

\bibitem[Lammert et~al.(2024)Lammert, Dreyer, Mathes, Kuligin, Borm, Schatz, Kiechle, L{\"o}rsch, Jung, Lange, et~al.]{lammert2024expert}
Jacqueline Lammert, Tobias Dreyer, Sonja Mathes, Leonid Kuligin, Kai~J Borm, Ulrich~A Schatz, Marion Kiechle, Alisa~M L{\"o}rsch, Johannes Jung, Sebastian Lange, et~al.
\newblock Expert-guided large language models for clinical decision support in precision oncology.
\newblock \emph{JCO precision oncology}, 8:\penalty0 e2400478, 2024.

\bibitem[Cohen(1960)]{cohen1960}
Jacob Cohen.
\newblock A coefficient of agreement for nominal scales.
\newblock \emph{Educational and Psychological Measurement}, 20\penalty0 (1):\penalty0 37--46, 1960.

\bibitem[Fleiss(1971)]{fleiss1971}
Joseph~L. Fleiss.
\newblock Measuring nominal scale agreement among many raters.
\newblock \emph{Psychological Bulletin}, 76\penalty0 (5):\penalty0 378--382, 1971.

\bibitem[Byrt et~al.(1993)Byrt, Bishop, and Carlin]{byrt1993}
Ted Byrt, Janet Bishop, and John~B. Carlin.
\newblock Bias, prevalence and kappa.
\newblock \emph{Journal of Clinical Epidemiology}, 46\penalty0 (5):\penalty0 423--429, 1993.

\bibitem[Gwet(2008)]{gwet2008}
Kilem~Li Gwet.
\newblock Computing inter-rater reliability and its variance in the presence of high agreement.
\newblock \emph{British Journal of Mathematical and Statistical Psychology}, 61\penalty0 (1):\penalty0 29--48, 2008.

\bibitem[Krippendorff(2011)]{krippendorff2011}
Klaus Krippendorff.
\newblock Computing krippendorff's alpha-reliability.
\newblock \url{https://repository.upenn.edu/asc_papers/43}, 2011.
\newblock University of Pennsylvania ScholarlyCommons.

\bibitem[Bankhead et~al.(2017)Bankhead, Loughrey, Fern{\'a}ndez, Dombrowski, McArt, Dunne, McQuaid, Gray, Murray, Coleman, et~al.]{bankhead2017qupath}
Peter Bankhead, Maurice~B Loughrey, Jos{\'e}~A Fern{\'a}ndez, Yvonne Dombrowski, Darragh~G McArt, Philip~D Dunne, Stephen McQuaid, Ronan~T Gray, Liam~J Murray, Helen~G Coleman, et~al.
\newblock Qupath: Open source software for digital pathology image analysis.
\newblock \emph{Scientific reports}, 7\penalty0 (1):\penalty0 1--7, 2017.

\bibitem[Vaidya et~al.(2025{\natexlab{b}})Vaidya, Zhang, Jaume, Song, Ding, Wagner, Lu, Doucet, Robertson, Almagro-Perez, et~al.]{vaidya2025molecular}
Anurag Vaidya, Andrew Zhang, Guillaume Jaume, Andrew~H Song, Tong Ding, Sophia~J Wagner, Ming~Y Lu, Paul Doucet, Harry Robertson, Cristina Almagro-Perez, et~al.
\newblock Molecular-driven foundation model for oncologic pathology.
\newblock \emph{arXiv preprint arXiv:2501.16652}, 2025{\natexlab{b}}.

\bibitem[Team et~al.(2025)Team, Kamath, Ferret, Pathak, Vieillard, Merhej, Perrin, Matejovicova, Ramé, Rivière, Rouillard, Mesnard, Cideron, bastien Grill, Ramos, Yvinec, Casbon, Pot, Penchev, Liu, Visin, Kenealy, Beyer, Zhai, Tsitsulin, Busa-Fekete, Feng, Sachdeva, Coleman, Gao, Mustafa, Barr, Parisotto, et~al.]{gemmateam2025gemma3technicalreport}
Gemma Team, Aishwarya Kamath, Johan Ferret, Shreya Pathak, Nino Vieillard, Ramona Merhej, Sarah Perrin, Tatiana Matejovicova, Alexandre Ramé, Morgane Rivière, Louis Rouillard, Thomas Mesnard, Geoffrey Cideron, Jean bastien Grill, Sabela Ramos, Edouard Yvinec, Michelle Casbon, Etienne Pot, Ivo Penchev, Gaël Liu, Francesco Visin, Kathleen Kenealy, Lucas Beyer, Xiaohai Zhai, Anton Tsitsulin, Robert Busa-Fekete, Alex Feng, Noveen Sachdeva, Benjamin Coleman, Yi~Gao, Basil Mustafa, Iain Barr, Emilio Parisotto, et~al.
\newblock Gemma 3 technical report.
\newblock \emph{arXiv preprint arXiv:2503.19786}, 2025.

\bibitem[OpenAI et~al.(2024)OpenAI, :, Hurst, Lerer, Goucher, Perelman, Ramesh, Clark, Ostrow, Welihinda, Hayes, Radford, Mądry, Baker-Whitcomb, Beutel, Borzunov, Carney, Chow, Kirillov, Nichol, Paino, Renzin, Passos, Kirillov, Christakis, Conneau, Kamali, Jabri, Moyer, Tam, Crookes, Tootoochian, Tootoonchian, Kumar, Vallone, Karpathy, Braunstein, Cann, et~al.]{openai2024gpt4ocard}
OpenAI, :, Aaron Hurst, Adam Lerer, Adam~P. Goucher, Adam Perelman, Aditya Ramesh, Aidan Clark, AJ~Ostrow, Akila Welihinda, Alan Hayes, Alec Radford, Aleksander Mądry, Alex Baker-Whitcomb, Alex Beutel, Alex Borzunov, Alex Carney, Alex Chow, Alex Kirillov, Alex Nichol, Alex Paino, Alex Renzin, Alex~Tachard Passos, Alexander Kirillov, Alexi Christakis, Alexis Conneau, Ali Kamali, Allan Jabri, Allison Moyer, Allison Tam, Amadou Crookes, Amin Tootoochian, Amin Tootoonchian, Ananya Kumar, Andrea Vallone, Andrej Karpathy, Andrew Braunstein, Andrew Cann, et~al.
\newblock Gpt-4o system card.
\newblock \emph{arXiv preprint arXiv:2410.21276}, 2024.

\bibitem[Zhu et~al.(2025)Zhu, Wang, Chen, Liu, Ye, Gu, Tian, Duan, Su, Shao, Gao, Cui, Wang, Cao, Liu, Wei, Zhang, Wang, Xu, Li, Wang, Deng, Li, He, Jiang, Luo, Wang, He, Shi, Zhang, Shao, He, Xiong, Qu, et~al.]{zhu2025internvl3exploringadvancedtraining}
Jinguo Zhu, Weiyun Wang, Zhe Chen, Zhaoyang Liu, Shenglong Ye, Lixin Gu, Hao Tian, Yuchen Duan, Weijie Su, Jie Shao, Zhangwei Gao, Erfei Cui, Xuehui Wang, Yue Cao, Yangzhou Liu, Xingguang Wei, Hongjie Zhang, Haomin Wang, Weiye Xu, Hao Li, Jiahao Wang, Nianchen Deng, Songze Li, Yinan He, Tan Jiang, Jiapeng Luo, Yi~Wang, Conghui He, Botian Shi, Xingcheng Zhang, Wenqi Shao, Junjun He, Yingtong Xiong, Wenwen Qu, et~al.
\newblock Internvl3: Exploring advanced training and test-time recipes for open-source multimodal models.
\newblock \emph{arXiv preprint arXiv:2504.10479}, 2025.

\bibitem[Grattafiori et~al.(2024)Grattafiori, Dubey, Jauhri, Pandey, Kadian, Al-Dahle, Letman, Mathur, Schelten, Vaughan, Yang, Fan, Goyal, Hartshorn, Yang, Mitra, Sravankumar, Korenev, Hinsvark, Rao, Zhang, Rodriguez, Gregerson, Spataru, Roziere, Biron, Tang, Chern, Caucheteux, Nayak, Bi, Marra, et~al.]{grattafiori2024llama3herdmodels}
Aaron Grattafiori, Abhimanyu Dubey, Abhinav Jauhri, Abhinav Pandey, Abhishek Kadian, Ahmad Al-Dahle, Aiesha Letman, Akhil Mathur, Alan Schelten, Alex Vaughan, Amy Yang, Angela Fan, Anirudh Goyal, Anthony Hartshorn, Aobo Yang, Archi Mitra, Archie Sravankumar, Artem Korenev, Arthur Hinsvark, Arun Rao, Aston Zhang, Aurelien Rodriguez, Austen Gregerson, Ava Spataru, Baptiste Roziere, Bethany Biron, Binh Tang, Bobbie Chern, Charlotte Caucheteux, Chaya Nayak, Chloe Bi, Chris Marra, et~al.
\newblock The llama 3 herd of models.
\newblock \emph{arXiv preprint arXiv:2407.21783}, 2024.

\bibitem[Bai et~al.(2025)Bai, Chen, Liu, Wang, Ge, Song, Dang, Wang, Wang, Tang, Zhong, Zhu, Yang, Li, Wan, Wang, Ding, Fu, Xu, Ye, Zhang, Xie, Cheng, Zhang, Yang, Xu, and Lin]{bai2025qwen25vltechnicalreport}
Shuai Bai, Keqin Chen, Xuejing Liu, Jialin Wang, Wenbin Ge, Sibo Song, Kai Dang, Peng Wang, Shijie Wang, Jun Tang, Humen Zhong, Yuanzhi Zhu, Mingkun Yang, Zhaohai Li, Jianqiang Wan, Pengfei Wang, Wei Ding, Zheren Fu, Yiheng Xu, Jiabo Ye, Xi~Zhang, Tianbao Xie, Zesen Cheng, Hang Zhang, Zhibo Yang, Haiyang Xu, and Junyang Lin.
\newblock Qwen2.5-vl technical report.
\newblock \emph{arXiv preprint arXiv:2502.13923}, 2025.

\bibitem[Yang et~al.(2025)Yang, Li, Yang, Zhang, Hui, Zheng, Yu, Gao, Huang, Lv, Zheng, Liu, Zhou, Huang, Hu, Ge, Wei, Lin, Tang, Yang, Tu, Zhang, Yang, Yang, Zhou, Zhou, Lin, Dang, Bao, Yang, Yu, Deng, Li, Xue, Li, Zhang, Wang, Zhu, Men, Gao, Liu, Luo, et~al.]{yang2025qwen3technicalreport}
An~Yang, Anfeng Li, Baosong Yang, Beichen Zhang, Binyuan Hui, Bo~Zheng, Bowen Yu, Chang Gao, Chengen Huang, Chenxu Lv, Chujie Zheng, Dayiheng Liu, Fan Zhou, Fei Huang, Feng Hu, Hao Ge, Haoran Wei, Huan Lin, Jialong Tang, Jian Yang, Jianhong Tu, Jianwei Zhang, Jianxin Yang, Jiaxi Yang, Jing Zhou, Jingren Zhou, Junyang Lin, Kai Dang, Keqin Bao, Kexin Yang, Le~Yu, Lianghao Deng, Mei Li, Mingfeng Xue, Mingze Li, Pei Zhang, Peng Wang, Qin Zhu, Rui Men, Ruize Gao, Shixuan Liu, Shuang Luo, et~al.
\newblock Qwen3 technical report.
\newblock \emph{arXiv preprint arXiv:2505.09388}, 2025.

\bibitem[Kwon et~al.(2023)Kwon, Li, Zhuang, Sheng, Zheng, Yu, Gonzalez, Zhang, and Stoica]{kwon2023efficient}
Woosuk Kwon, Zhuohan Li, Siyuan Zhuang, Ying Sheng, Lianmin Zheng, Cody~Hao Yu, Joseph~E. Gonzalez, Hao Zhang, and Ion Stoica.
\newblock Efficient memory management for large language model serving with pagedattention.
\newblock In \emph{Proceedings of the ACM SIGOPS 29th Symposium on Operating Systems Principles}, 2023.

\end{thebibliography}

\clearpage

\appendix
\newpage

\section*{\Large Appendix}

\section{Background on Molecular Tumor Boards} \label{sec:mtb}

\begin{figure}[h!]
    \centering
    \includegraphics[width=0.99\linewidth]{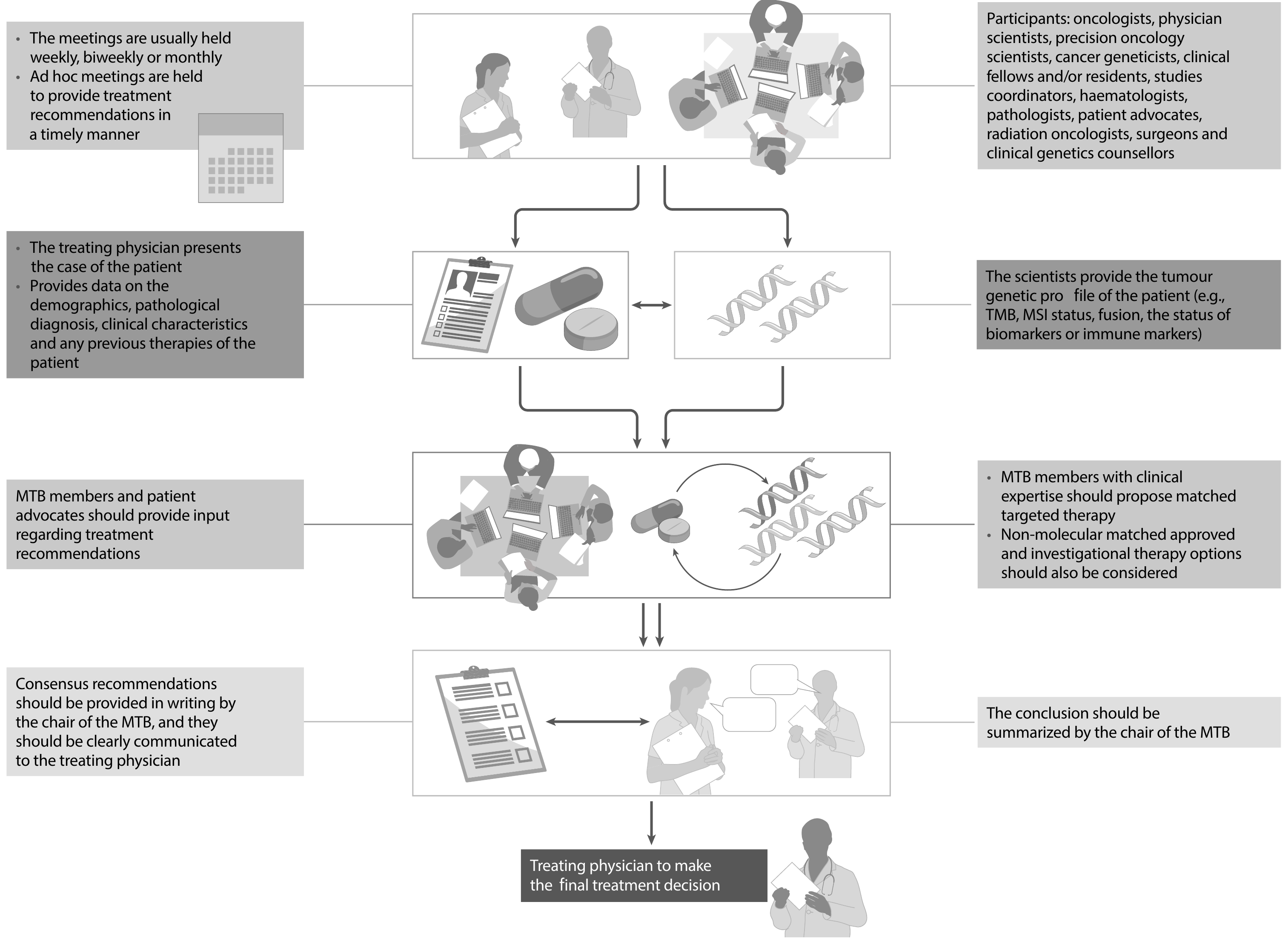}
    \caption{\footnotesize \textbf{Overview of the Molecular Tumor Board process.} Meetings are held at regular intervals or on demand, bringing together multidisciplinary experts who jointly review patient history, molecular profiling results, and clinical evidence to recommend personalized treatment strategies. Final decisions are communicated in writing to the treating physician. Figure adapted from \citet{Tsimberidou2023}.}
    \label{fig:mtb}
\end{figure}

Molecular Tumor Boards are structured, multidisciplinary forums where complex cancer cases are evaluated through the integration of clinical, pathological, and molecular data. These boards include oncologists, pathologists, geneticists, and other specialists who collectively interpret diagnostic and genomic findings to formulate personalized treatment recommendations. The process begins with the treating physician presenting the patient’s case, including demographic and clinical background, prior therapies, and pathological findings. Scientists contribute molecular profiling data—such as tumor mutational burden (TMB), microsatellite instability (MSI) status, fusion events, and biomarker expression—that are critical for matching patients to targeted therapies \citep{Tsimberidou2023}.

Treatment options are proposed by the MTB members based on this integrated evidence, and consensus recommendations are recorded and relayed to the treating physician for final decision-making. The process is iterative, often revisiting cases as new data becomes available, and increasingly involves patient advocates and real-time access to clinical trial data and treatment databases.

Figure~\ref{fig:mtb} visualizes this workflow, emphasizing how the flow of structured information and expert input leads to treatment recommendations that are tailored, evidence-driven, and context-aware. This structured, evolving nature of clinical decision-making underpins the \our benchmark design, which mirrors MTB dynamics through sequential, multimodal question-answering tasks grounded in expert-reviewed patient cases.

\section{Related Work} \label{sec:related_works}

\paragraph{Benchmarks for clinical AI}
Numerous benchmarks have been developed to evaluate clinical question-answering and medical reasoning in large language models. Early efforts focused on textual data, such as EHRNoteQA~\citep{kweon2024ehrnoteqa}, which uses discharge summaries to evaluate clinical understanding, and MedQA~\citep{medqa}, which tests medical licensing exam-style questions. While these benchmarks provide foundational testbeds, they are largely unimodal and do not evaluate temporal reasoning or interaction. ClinicBench~\citep{Liu2024ClinicBench} and MedJourney~\citep{medjourney} introduced more realistic clinical tasks with timeline structures, but still lack multimodal integration and agentic file access. \our addresses these gaps by combining longitudinal structure, multimodal data, and interactive information retrieval within a single benchmark.

\paragraph{Multimodal biomedical benchmarks}
Multimodal benchmarks such as MC-BEC~\citep{chen_multimodal_2023}, Asclepius~\citep{wang2024asclepius}, GMAI-MMBench~\citep{chen2024gmaimmbenchcomprehensivemultimodalevaluation}, and MedTrinity-25M~\citep{xie2024medtrinity25mlargescalemultimodaldataset} have broadened evaluation to include clinical text, waveforms, and images, testing foundation models across diverse modalities. However, these tasks are typically static and do not assess models’ ability to retrieve, interpret, or combine information across time. Pathology-specific datasets like HEST-1k~\citep{jaume2024hest} and PANDA~\citep{panda} enable visual classification, but lack the clinical reasoning components required for decision-making. \our advances this line of work by evaluating whether models can synthesize insights across pathology, hematology, and genomics within evolving case contexts.

\paragraph{Agentic clinical systems}
 Recent interest in LLM-based clinical agents has led to new frameworks such as MedAgentBench~\citep{jiang2025medagentbenchrealisticvirtualehr}, MediQ~\citep{li2024mediqquestionaskingllmsbenchmark}, MedAgent-Pro~\citep{wang2025medagentproevidencebasedmultimodalmedical}, RadA-BenchPlat~\citep{zheng2025modernllmsactagent}, and AgentClinic~\citep{schmidgall2024agentclinicmultimodalagentbenchmark}. These systems evaluate agents in interactive or dialogue-based environments, but focus primarily on textual data or synthetic tasks. There has been effort to develop LLM-based agents for chest X-rays~\citep{fallahpour2025medraxmedicalreasoningagent}, histopathology~\citep{ghezloo2025pathfinder}, and multiple imaging modalities~\citep{wang2025medagent}.  \our complements these efforts by embedding tool-use into clinically realistic workflows and measuring how tool-augmented agents reason in complex, multimodal scenarios.

\paragraph{Foundation models in healthcare}
Foundation models trained on biomedical corpora or medical imaging datasets have demonstrated promising capabilities in generalization and zero-shot reasoning~\citep{moor2023foundation}. Biomedical language models trained on literature and structured data (e.g., BioGPT~\citep{10.1093/bib/bbac409}, BioMedLM~\citep{bolton2024biomedlm27bparameterlanguage}) support evidence grounding. In pathology, several vision foundation models have emerged: H-optimus-0~\citep{hoptimus0}, Phikon-v2~\citep{filiot2024phikonv2largepublicfeature}, Virchow~\citep{Vorontsov2024}, and PathOrchestra~\citep{yan2025pathorchestracomprehensivefoundationmodel}. Vision-language models like CONCH~\citep{CONCH}, UNI-2~\citep{chen2024uni}, and MUSK~\citep{Xiang2025} integrate histopathology images with clinical text to enhance slide interpretation and support precision oncology applications. However, these models are typically benchmarked in isolation. \our instead evaluates how foundation models function as tools within agentic systems, testing not just their predictive accuracy but also their integration into sequential decision-making processes.

\paragraph{Precision oncology and Molecular Tumor Boards}
MTBs represent a high-stakes, information-dense setting in which multimodal and longitudinal reasoning is essential~\citep{Tsimberidou2023}. \citet{lammert2024expert} introduces a domain-specific LLM system for oncology treatment recommendations. Prior datasets in this space are often limited to structured formats or single-modality use cases. \our is among the first benchmarks to simulate MTB workflows comprehensively, capturing the clinical sequencing, agent interaction, and data integration that define real-world oncology decision-making~\citep{msk}.

\section{Clinical Validation}
\subsection{Companion App for Clinical Validation} \label{sec:companion_app}

\begin{figure}[h!]
    \centering
    \includegraphics[width=\linewidth]{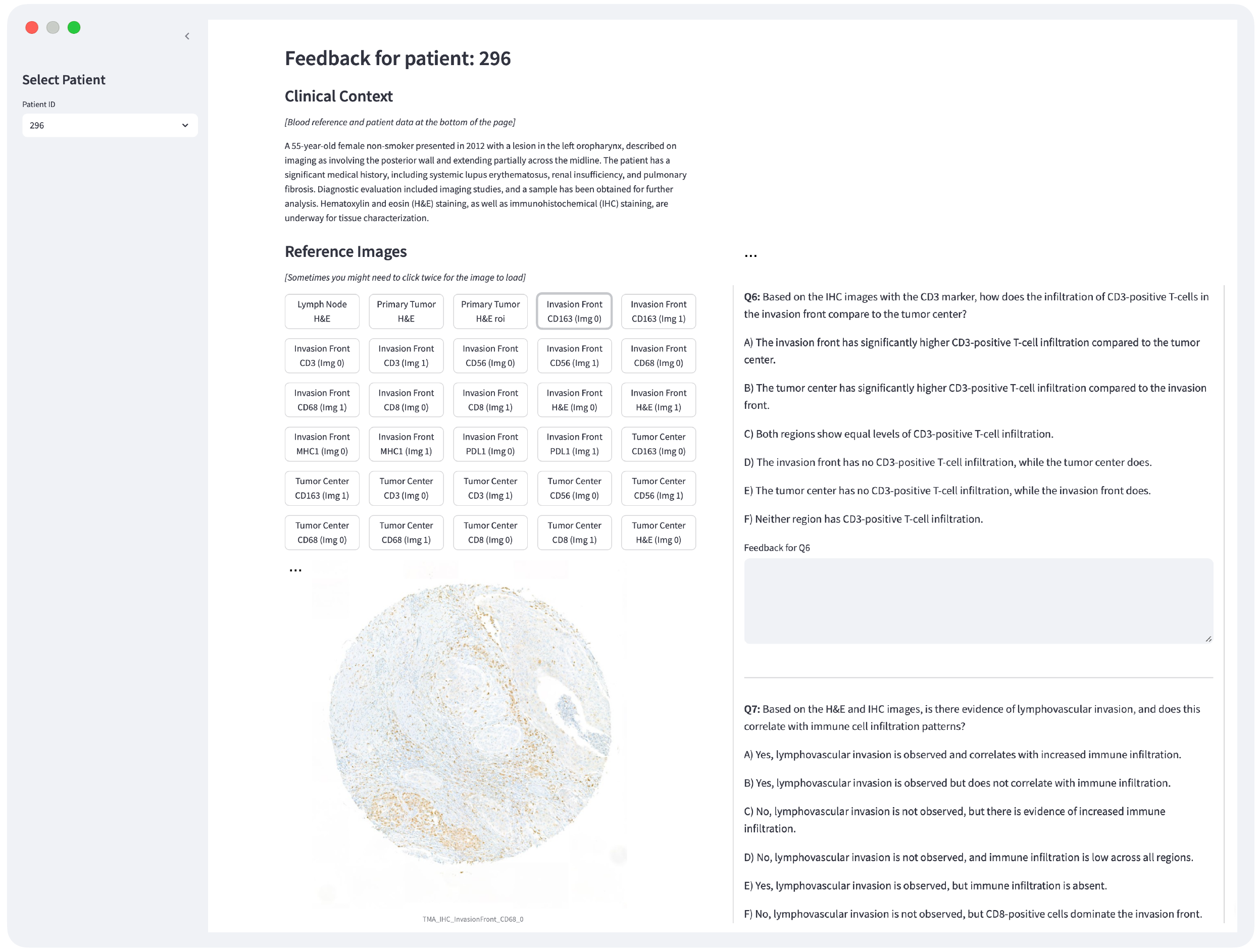}
    \caption{\footnotesize \textbf{Companion app interface for clinical validation.} The platform displays clinical context, reference images grouped by region and marker, and multiple-choice questions for expert review. Full-resolution slide viewers and inline feedback fields allow for efficient validation of benchmark items.}
    \label{fig:comanion_app}
\end{figure}

To ensure the clinical validity of the benchmark and facilitate expert-in-the-loop review, we developed a custom web-based application to support question and answer annotation (\url{https://share.streamlit.io/app/oncoform/} - private). The companion app allows clinicians to browse detailed patient cases, including demographic data, clinical summaries, reference images (H\&E and IHC), and the associated question-answer pairs (Fig.~\ref{fig:comanion_app}).

Each patient case is presented with a structured clinical context, followed by a grid of labeled image thumbnails categorized by tissue region and marker (e.g., "Tumor Center CD8", "Invasion Front H\&E"). Clicking a thumbnail loads the full-resolution slide. Besides the image panel, the user is presented with multiple-choice questions linked to the case, and input fields to provide feedback or corrections for each answer.

This interface was designed to mirror the decision-making process of molecular tumor boards, offering domain experts an intuitive environment for validating multimodal reasoning tasks. It supports both qualitative feedback and quantitative validation, and was used extensively during benchmark construction to curate expert-reviewed question sets for both \our-Multimodal and \our-Longitudinal.

\subsection{Expert Validation}

After multiple iterations of internal expert-in-the-loop auditing, the final set of questions in \our{} were sent for external manual review by domain specialists via the companion application, where we aimed to judge the soundness and relevance of the QA-pairs. Due to the diverse nature of the benchmark and different categories of questions, reviews were conducted both independently and with overlapping sets to assess consistency. In total, 10 experts from multiple countries, hospitals, and levels of expertise were involved in the review process. For the overlapping sets, the external reviewers received 45 questions to review via the app, and the QA pairs were rated as Good as is, Needs minor rewording, or Inappropriate. We report the average inter-rater agreement metrics over these reviews in Table~\ref{tab:agreement}. We observe that there is a high annotation consistency and confidence in question quality. 

\begin{table}[htbp!]
\centering
\caption{Average inter-rater agreement metrics of ten experts over 45 questions.}
\begin{adjustbox}{max width=\textwidth}
\begin{tabular}{lccccccc}
\toprule
Overall & Mean pairwise & Mean Cohen's $\kappa$ \citep{cohen1960} & Fleiss' $\kappa$ \citep{fleiss1971} & Mean PABAK \citep{byrt1993} & Gwet's AC1 \citep{gwet2008} & Krippendorff's $\alpha$ \citep{krippendorff2011} \\
\midrule
0.91 & 0.94 & 0.81 & 0.79 & 0.91 & 0.91 & 0.79 \\
\bottomrule
\end{tabular}
\end{adjustbox}
\label{tab:agreement}
\end{table}

\section{Additional Details on \our}

\subsection{Agentic Workflow} \label{sec:agentic_workflow}

\paragraph{Interactive dialogue setup} To simulate realistic clinical scenarios, \our employs an interactive, multi-turn dialogue setup in which a Doctor agent is tasked with responding to several user questions. These questions are grounded in access to several modalities grouped in a comprehensive patient case folder.

\paragraph{Conversation initialization} Both evaluation datasets follow the same general approach, where the conversation begins with a short patient introduction, which contains patient demographic information such as age, gender, as well as symptoms the patient is currently experiencing about their oncological condition. The agent is then guided on the availability and structure of files that contain additional patient-specific information. 

\paragraph{Differences between datasets} The available files vary between the two \our datasets. In the multimodal dataset, patient data can include digital pathology, hematological, and general clinical notes. In contrast, the longitudinal dataset presents information in temporal or tabular format, reflecting the patient's medical history over time.

\paragraph{Agent actions and memory constraints} The agent is equipped with two main actions during its reasoning: it can either request one or several files or provide a final answer once it has gathered sufficient information. Files accessed by the agent remain available only during the context of the current question being answered. For instance, if the model receives several images or text files while responding to a query, those files will no longer be accessible when answering subsequent questions, unless the model requests them again. This setup simulates a realistic clinical workflow, where an agent opens several clinical files from a patient folder, processes their information, and closes them upon completion of the current task. As a result, only the agent's reasoning steps alongside the record of which files were accessed persist across turns, reinforcing the need for information gathering and memory management.

\paragraph{Progressive contextual disclosure} As the conversation progresses, with multiple questions addressed, new contextual information and corresponding patient files are made available to the agent. This design mirrors a real-life clinical workflow, where the initial diagnosis is performed, and several additional tests are performed over time. By introducing new stages, the agent is required to continuously reason over evolving patient information.

\paragraph{\our-Multimodal structure} In the multi-modal part of our benchmark, the conversation begins with initial patient information, accompanied by multiple H\&E and IHC slides, provided to support reasoning around oncological image interpretation. With the initial assessment performed and primary cancer type identified, the simulated environment continues with a pre-surgical stage. In preparation for a surgery, the patient undergoes several lab tests aimed at assessing the overall health status and fitness for surgery and future treatments. In this stage, the agent has access to the patient's blood tests, together with a reference table specifying normal ranges for male and female patients. Following this, the surgery would be carried out, and a summary of the outcome would be provided in the context. The full surgery report and a short list of interventions would be accessible in the patient's case folder. At this stage, the questions asked would have a prognostic nature, with the agent tasked to predict the 5-year survival outcome and 2-year cancer recurrence, based on the patient data gathered throughout the entire case.

\paragraph{\our-Longitudinal structure} In the longitudinal evaluation track, the conversation begins with baseline clinical information and the introduction of the primary cancer type. The questions asked target outcome prediction, recurrence, and cancer progression for a given period. To support the agent in its reasoning, a timeline file is available in the patient's case folder with important events sorted by the age at which they occurred. If a sample has been taken and sequenced, additional patient data is provided in tabular format, such as gene mutations, copy number alterations, and structural variants. After the agent has answered several questions, a new context may be provided containing the outcome of all questions asked, with a new timeline file capturing the additional patient history. The agent can then combine all timeline files together to create a complete and comprehensive medical history to answer additional questions.

\paragraph{Evaluation and reproducibility} To facilitate reproducibility and streamline model evaluation, each run stores the complete conversation history, including the model's final answers for each question, the set of files accessed per query, and any hallucinated file names. We provide comprehensive logs for all evaluated models, both with and without tool access, in the project repository.

\section{Details on Experiments} \label{sec:exp_details}

\subsection{Details on Foundation Models and Downstream Tasks} \label{sec:abmil}

\paragraph{H\&E foundation model}
For histopathology image encoding, we employ the CONCH model directly, without any task-specific fine-tuning on H\&E slides. Model weights are obtained from the \href{https://huggingface.co/MahmoodLab/CONCH}{official HuggingFace repository}. To ensure compatibility and optimal performance, we adhere strictly to the preprocessing and usage guidelines outlined in the model card, including image normalization and text-token preparation for the dual-encoder architecture.

\paragraph{CONCH downstream tasks}
The CONCH foundation model is utilized for zero-shot region-of-interest (ROI) classification, where candidate labels are supplied by the LLM. Classification is performed by computing the dot product similarity between image and text embeddings. The label with the highest similarity is selected and returned to the LLM, along with a confidence estimate. To improve interpretability and account for potential label ambiguity, the raw softmax score is discretized into confidence bins: \textit{very low} (0–20\%), \textit{low} (20–40\%), \textit{medium} (40–60\%), \textit{high} (60–80\%), and \textit{very high} (80–100\%). Because the LLM-generated label set may not always include the most semantically accurate class, we avoid exposing exact probabilities and instead provide only the corresponding bin.

\paragraph{Tissue extraction using QuPath \citep{bankhead2017qupath}} To develop our ABMIL tool, which predicts the percentage of positively stained cells, we manually curated a dataset of immunohistochemistry (IHC) images using a semi-automated annotation pipeline in QuPath. For each patient, tissue microarray (TMA) cores were segmented to isolate regions containing tissue, and each region was mapped to the corresponding patient metadata. Within these regions, individual cells were identified through a combination of optical density transformation, background correction, and morphological segmentation. Cells were then classified as positively or negatively-stained based on DAB staining intensity. The resulting cell-level annotations formed the basis for training and evaluating our ABMIL model.

\paragraph{IHC foundation model} We extract UNIv2 embeddings using the TRIDENT framework \citep{vaidya2025molecular}. Each whole-slide image is first loaded at the appropriate microns-per-pixel (mpp) resolution, followed by tissue segmentation using the HEST model \citep{jaume2024hest}. The segmented tissue regions are then divided into fixed-size patches of $256 \times 256$ pixels. The UniV2 foundation model is applied to each patch to generate feature embeddings, which are subsequently aggregated using the ABMIL.

Our Gated ABMIL model architecture consists of two attention heads, each with a dimensionality of 512. The model incorporates a dropout rate of 0.3 and an overall model dimensionality of 1536. The regression component of the IHC tool is implemented as a five-layer fully connected neural network, utilizing ReLU activation functions and a dropout rate of 0.2. Training was conducted over 70 epochs using a batch size of 64, the Adam optimizer, and a learning rate of 0.0004. All experiments were carried out on a single NVIDIA A100 GPU with 80 GB of memory.

We release the pretrained weights of our IHC foundation model as part of the project’s GitHub repository. 

\subsection{Details on Resource and Knowledge Database Tools} \label{sec:classical_tools}

\paragraph{PubMed querying tool}
To retrieve biomedical literature, we integrate a PubMed querying tool using the Biopython library (\url{https://biopython.org/}). Search queries are generated by the calling LLM and may include advanced formatting, such as boolean operators (e.g., \texttt{lung carcinoma OR lung adenocarcinoma}). The tool retrieves the top 30 articles from PubMed based on the provided query and reranks them using the \texttt{BAAI-bge-reranker-v2-m3} model. The reranker jointly embeds the query and each article's abstract as a text pair and computes a logit-based relevance score. The top 3 abstracts with the highest scores are returned to the LLM for downstream reasoning.

\paragraph{DrugBank querying tool}
To integrate drug-related knowledge, we obtained a non-commercial research license for DrugBank and accessed the platform’s API. We retrieved a snapshot containing approximately 21,000 drug names along with their corresponding descriptions. Our DrugBank querying tool performs string-based lookups on file contents requested by the LLM. When a drug name match is identified, the corresponding description is returned to the LLM, enabling enhanced contextual understanding and more informed clinical reasoning.

\subsection{Details on Large Language and Vision Language Models}  

\paragraph{LLM specifications}
We benchmark the following models: \gemma{12}, \gemma{27} \citep{gemmateam2025gemma3technicalreport}, \gpt\ \citep{openai2024gpt4ocard}, \href{https://openai.com/index/introducing-o3-and-o4-mini/}{o4-mini}, \internvl{38}, \internvl{78} \citep{zhu2025internvl3exploringadvancedtraining}, \llamavision{90}, \llamaold{8}, \llama{70} \citep{grattafiori2024llama3herdmodels}, \href{https://mistral.ai/news/mistral-small-3-1}{\mistralsmall{24}}, \qwenvl{7}, \qwenvl{32} \citep{bai2025qwen25vltechnicalreport}, \qwennew{8}, \qwennew{32} \citep{yang2025qwen3technicalreport}. 
To optimize GPU memory usage while preserving model performance, we apply 4-bit quantization to the following models: \llamavision{90}, \mistralsmall{24}, \llamaold{8}, \llama{70}, \qwennew{8} (reasoning), and \qwennew{32} (reasoning). These models are served using the VLLM inference engine \citep{kwon2023efficient}, with weights sourced from HuggingFace. For \gemma{12}, \gemma{27}, \qwenvl{7}, and \qwenvl{32}, we employ 8-bit quantization and use the HuggingFace implementation and associated pretrained weights. For OpenAI-based models, we use the \texttt{gpt-4o-2024-08-06} checkpoint for \gpt{} and the \texttt{o4-mini-2025-04-16} checkpoint for \ofourmini{}.

\subsection{Details on Evaluation Metrics}

\paragraph{Answer accuracy}
We evaluate the agent system primarily using accuracy, measured on a set of true/false and multiple-choice questions (each with six answer options). This formulation enables objective evaluation without relying on human annotators or oracle LLMs, thereby ensuring reproducibility and consistency across models. Model outputs are parsed using regular expressions to extract answers. We attempt to identify whether the output is a single letter (e.g., \texttt{[ANSWER: A]}) or a letter with the corresponding option (e.g., \texttt{[ANSWER: A) Squamous Cell Carcinoma, Keratizing]}). If the model does not follow this format, we prompt it again to extract a valid answer. After three failed attempts, the response is marked as incorrect.

\paragraph{File access count}
In addition to accuracy, we track the number of files accessed per question. The model is allowed to access files by explicitly specifying the filename and extension in the prescribed format (\ie, \texttt{[REQUEST: primary\_tumor\_roi.jpg]}). This metric captures how actively the model explores the available patient data, serving as a proxy for information-seeking behavior. A lower file access count may indicate superficial reasoning or hallucination, whereas higher counts suggest information retrieval and more grounded decision-making. Thus, this metric provides valuable insight into the agent's interpretability and alignment with real-world clinical workflows. 

\subsection{Details on Computational Resources}

All agentic experiments were conducted using NVIDIA A100 80GB GPUs. Specifically, the models \qwennew{8} (reasoning), \gemma{12}, \llamaold{8}, and \qwenvl{7} were run on a single GPU. We used two GPUs for \gemma{27}, \qwenvl{32}, \qwennew{32} (reasoning), and \mistralsmall{24}. The most resource-intensive model, \llama{70}, required seven GPUs for inference. For the OpenAI models \gpt{} and \ofourmini{}, all evaluations were performed using CPU-only inference via API access. Across all configurations, the full execution of the \our benchmark for a given model was completed within a maximum of two days.

\newpage

\begin{minipage}{\textwidth}
\section{Examples of \our Agentic Interactions}
\label{sec:examples}
    \centering
    \includegraphics[width=0.99\linewidth]{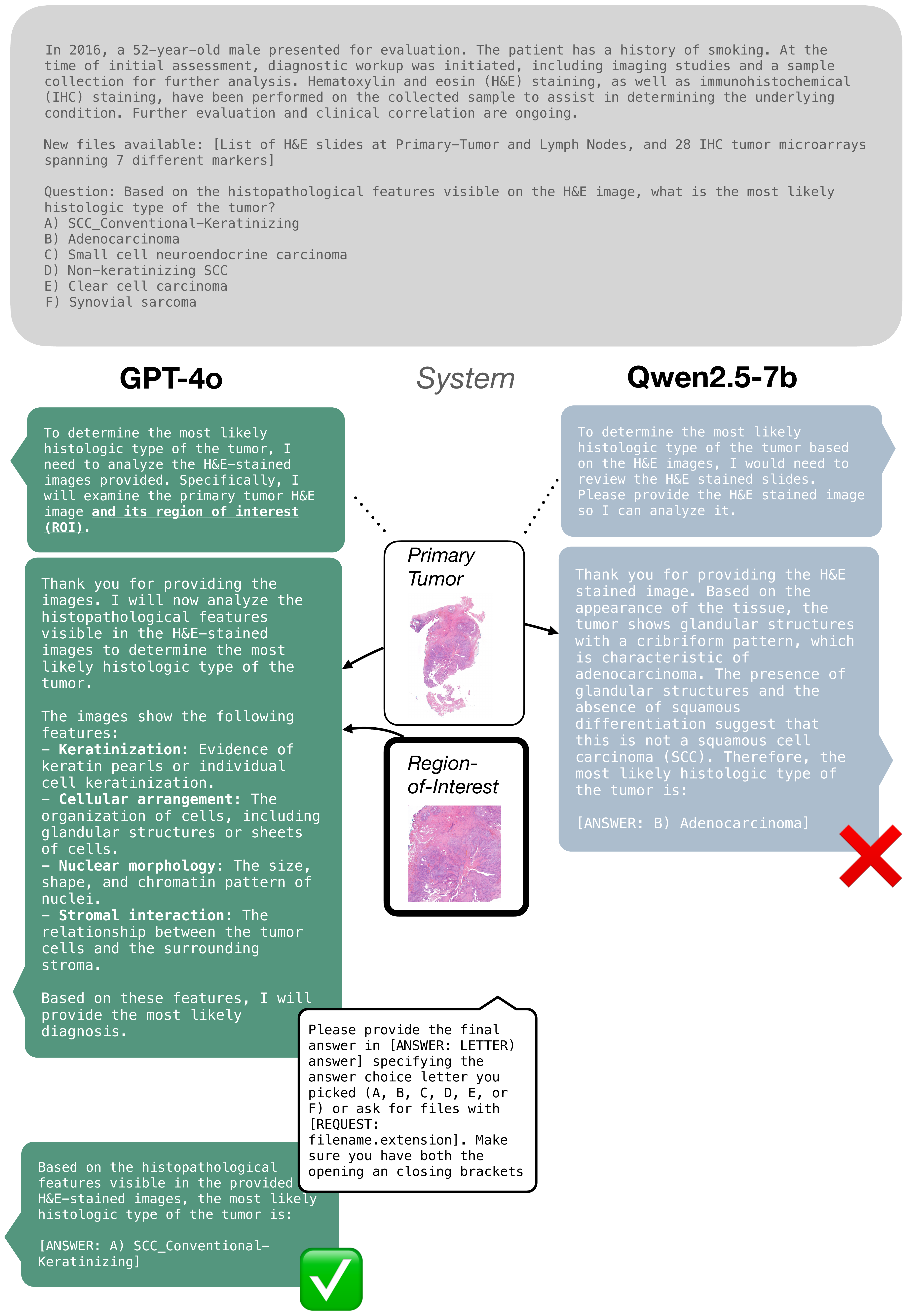}
    \captionof{exmpl}{\footnotesize In this digital pathology question, \gpt\ requests not only the low-resolution primary tumor slide, but also the high resolution region-of-interest (ROI), which allows to identify the correct cancer subtype. On the other hand, \qwenvl{7} cannot identify the correct subtype because it only requested the primary tumor slide, which is too high level to identify the critical features for cancer subtype identification. Note that the raw histology slides are high resolution (which allows to extract ROIs) but were resized for compatibility with VLMs.}
    \label{fig:example_1}
\end{minipage}

\begin{minipage}{\textwidth}
    \includegraphics[width=0.99\linewidth]{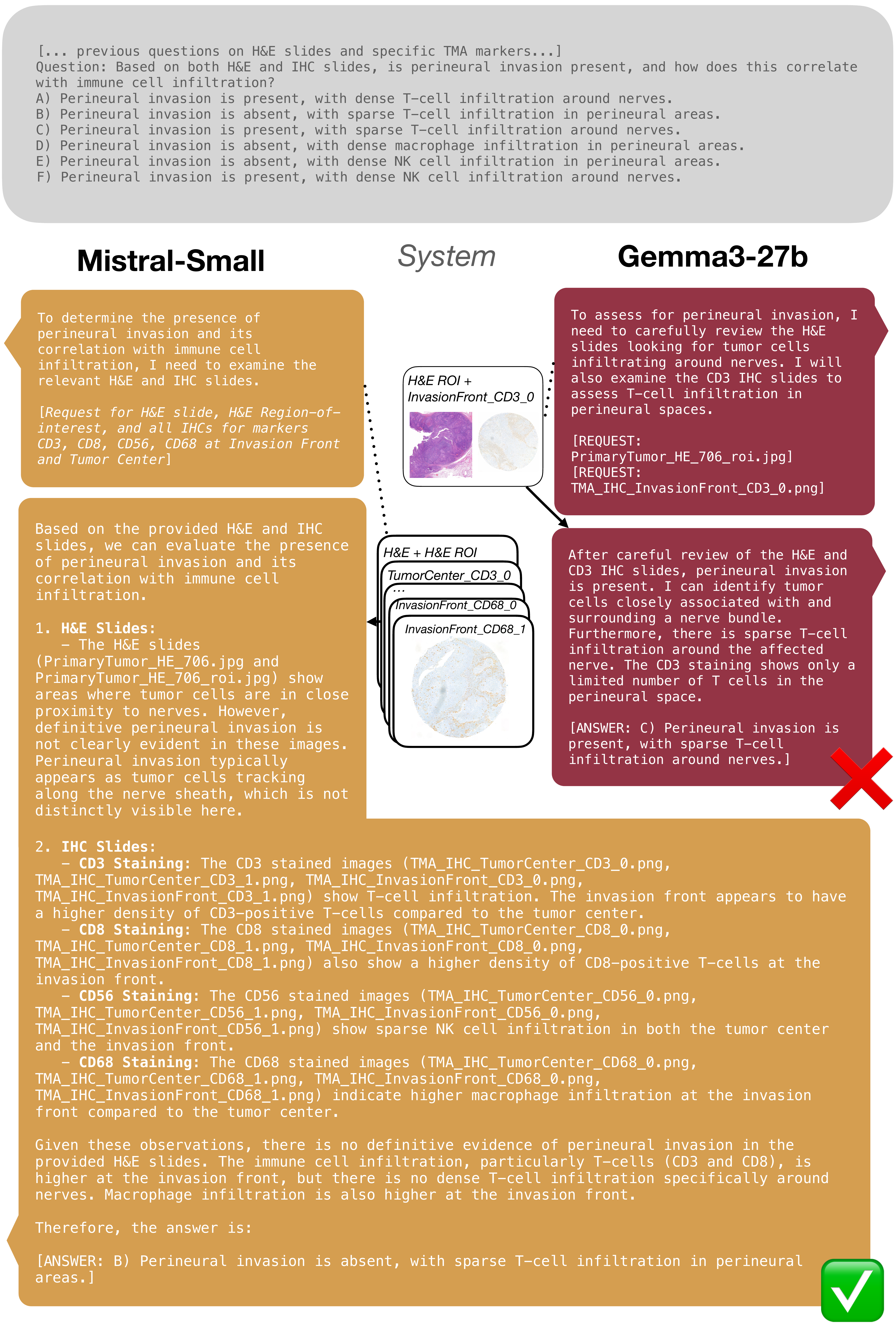}
   \captionof{exmpl}{\footnotesize In this digital pathology question, \mistralsmall{24} requests the histology slide and ROI, as well as 16 IHC images corresponding to two images for each of 4 markers observed at Tumor Center and Invasion Front. On the other hand, \gemma{27} only requests the histology ROI and one IHC, which leads to a wrong analysis.}
    \label{fig:example_2}
\end{minipage}

\begin{exmpl}[H]
    \centering
    \includegraphics[width=0.99\linewidth]{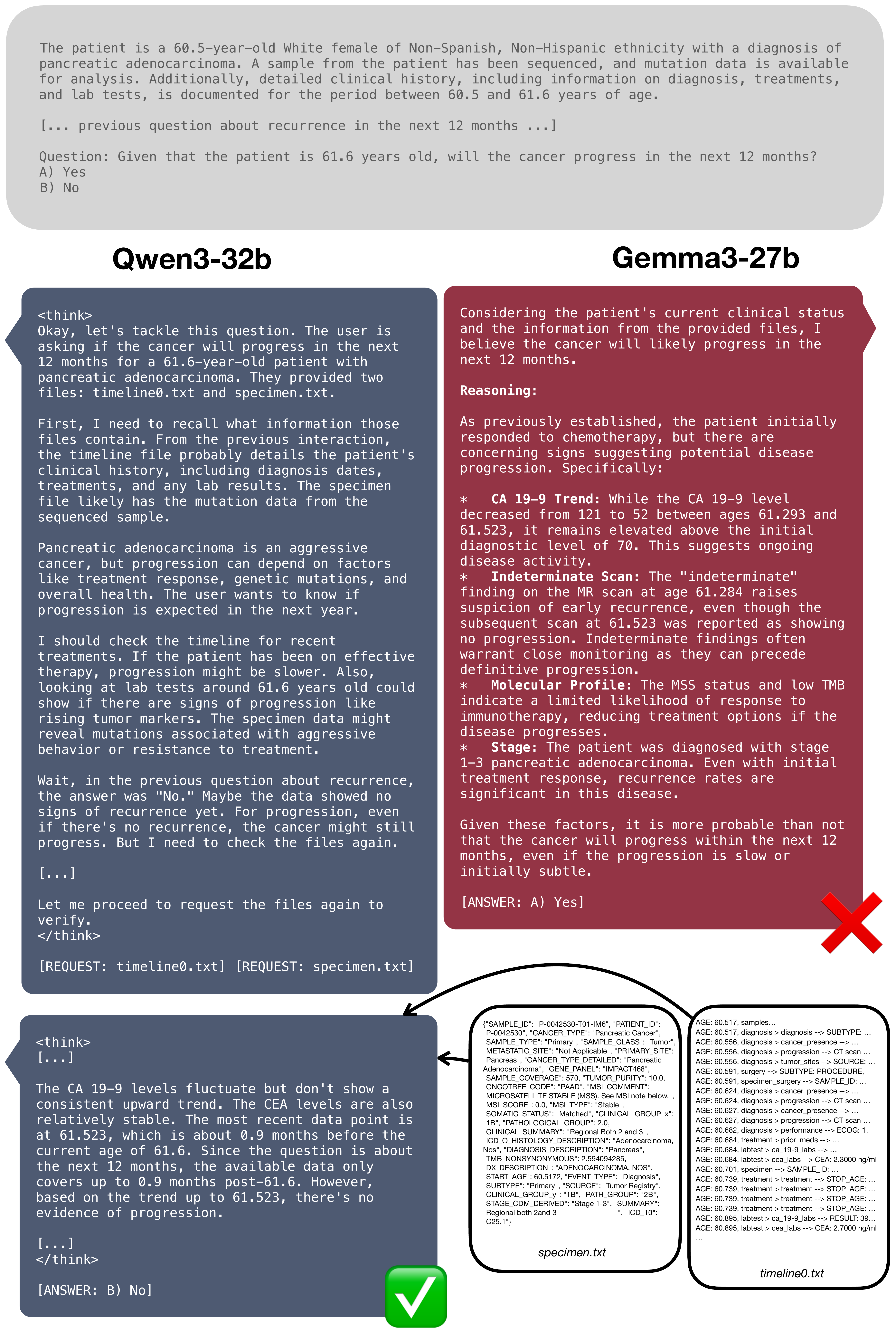}
    \caption{\footnotesize In this longitudinal case, the question asks about cancer \textit{progression} in the next months. However, previous questions already asked about cancer \textit{recurrence}, so \gemma{27} re-uses the data analysis from the previous question, without requesting for the timeline data or pathological data again. On the other hand, \qwennew{32} reasons that \textit{progression} may not require the same analysis than \textit{recurrence} and requests for the files again, which leads to the correct analysis.}
    \label{fig:example_3}
\end{exmpl}

\begin{exmpl}[h!]
    \centering
    \includegraphics[width=0.99\linewidth]{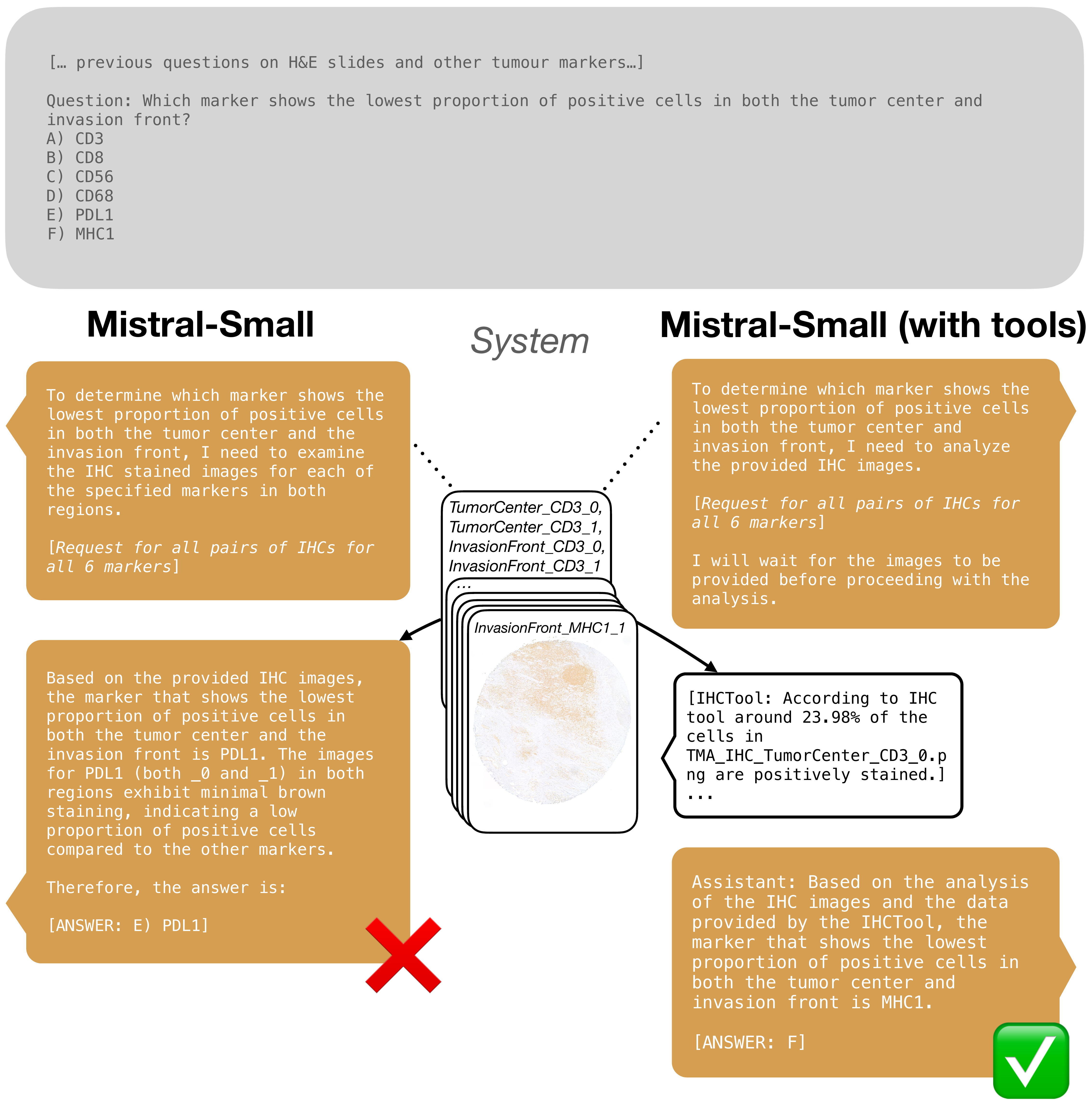}
    \caption{\footnotesize Augmenting LLMs with tools such as foundation models allows for grounded and explainable predictions. In this case, the tool-augmented model uses the output of the ABMIL model that quantifies the percent of positive cells in a IHC core, which, in contrast to the zero-shot LLM, leads to the correct answer. }
    \label{fig:example_4}
\end{exmpl}


\newpage

\section{Further Experimental Results} \label{sec:further_results}

\begin{figure}[H]
    \centering
    \includegraphics[width=0.99\linewidth]{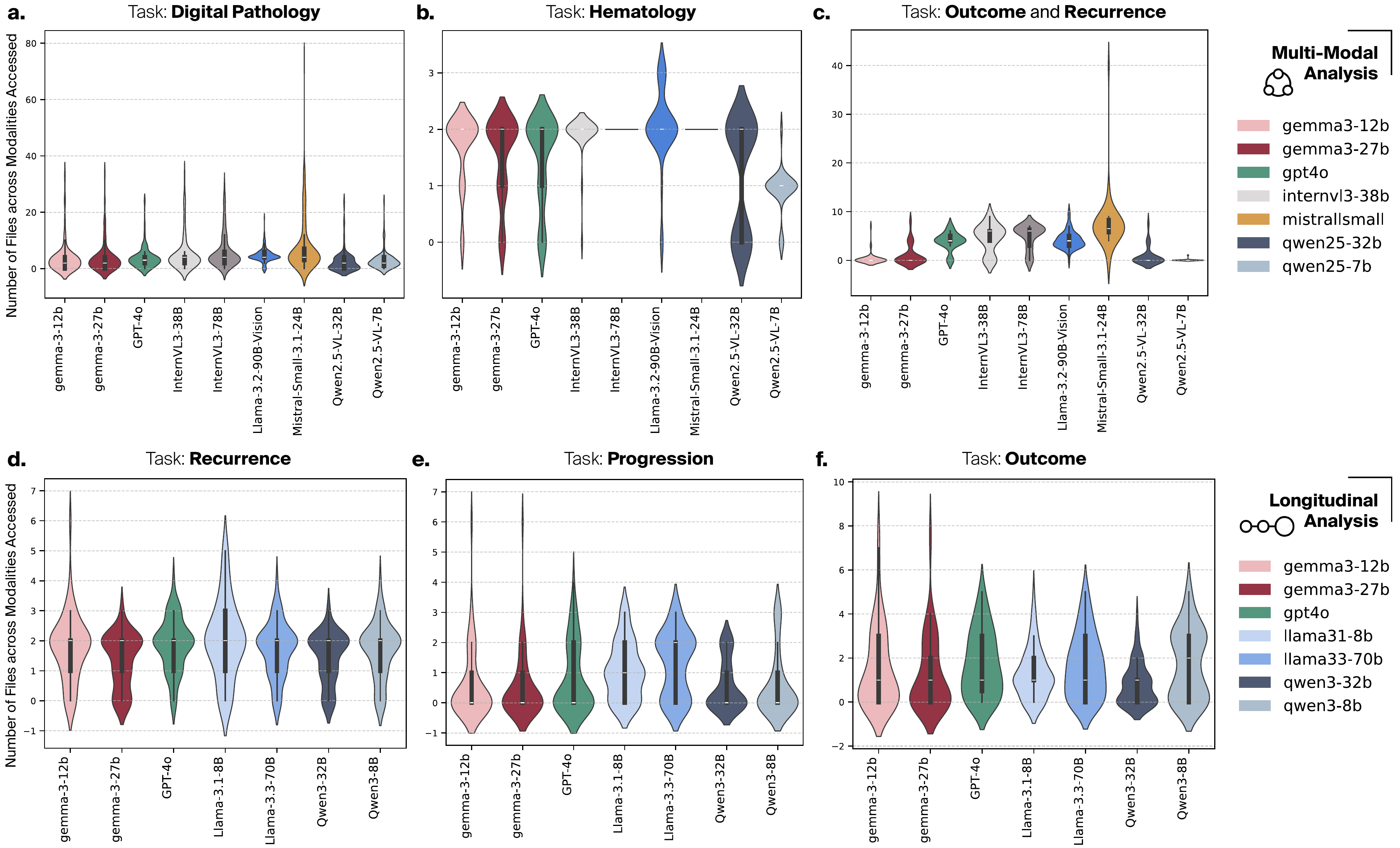}
    \caption{\footnotesize Distribution of number of files requested for a given backbone LLM per task.}
    \label{fig:violins}
\end{figure}

\begin{figure}[h]
    \centering
    \includegraphics[width=0.99\linewidth]{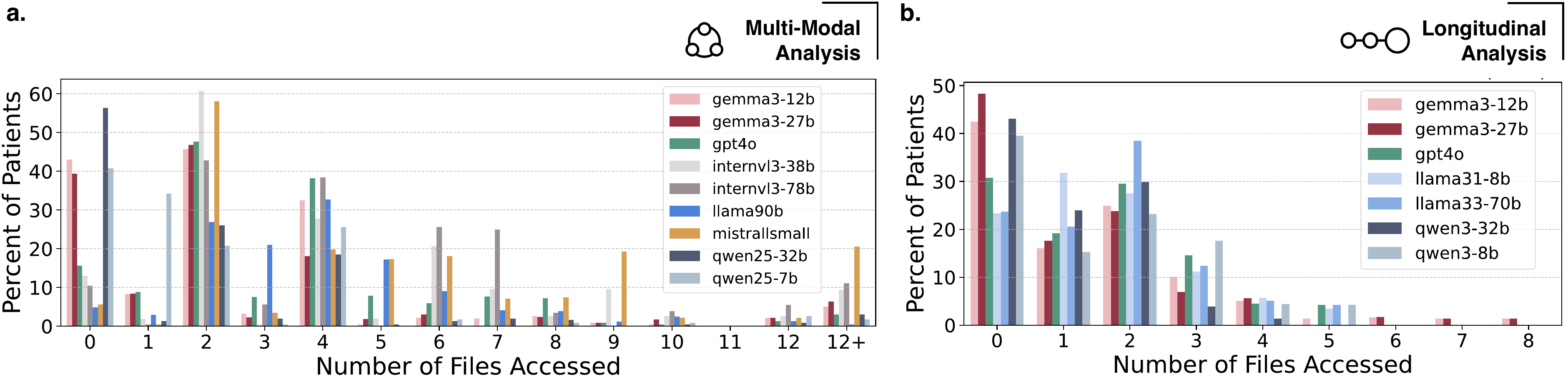}
    \caption{\footnotesize \textbf{Distribution of the number of files accessed per patient.} Across models for \our-Multimodal \textbf{a.} and \our-Longitudinal \textbf{b.} tasks.}
    \label{fig:pc_patients_num_files}
\end{figure}

Figure~\ref{fig:violins} shows the distribution of requested files per model and task. Some tasks show consistent file requests across models: for example, hematology usually requires two files to answer questions, namely patient hematology data and hematology reference ranges. However, some harder tasks like Outcome \& Recurrence show more variability across models. For example, \mistralsmall{24} tends to ask for more files than any other model, which could explain how it reaches a similar accuracy in this task as \llamavision{90}, a model that contains almost four times as many parameters (see Fig.~\ref{fig:combined_dots}). 

\begin{figure}[h]
    \centering
    \includegraphics[width=0.99\linewidth]{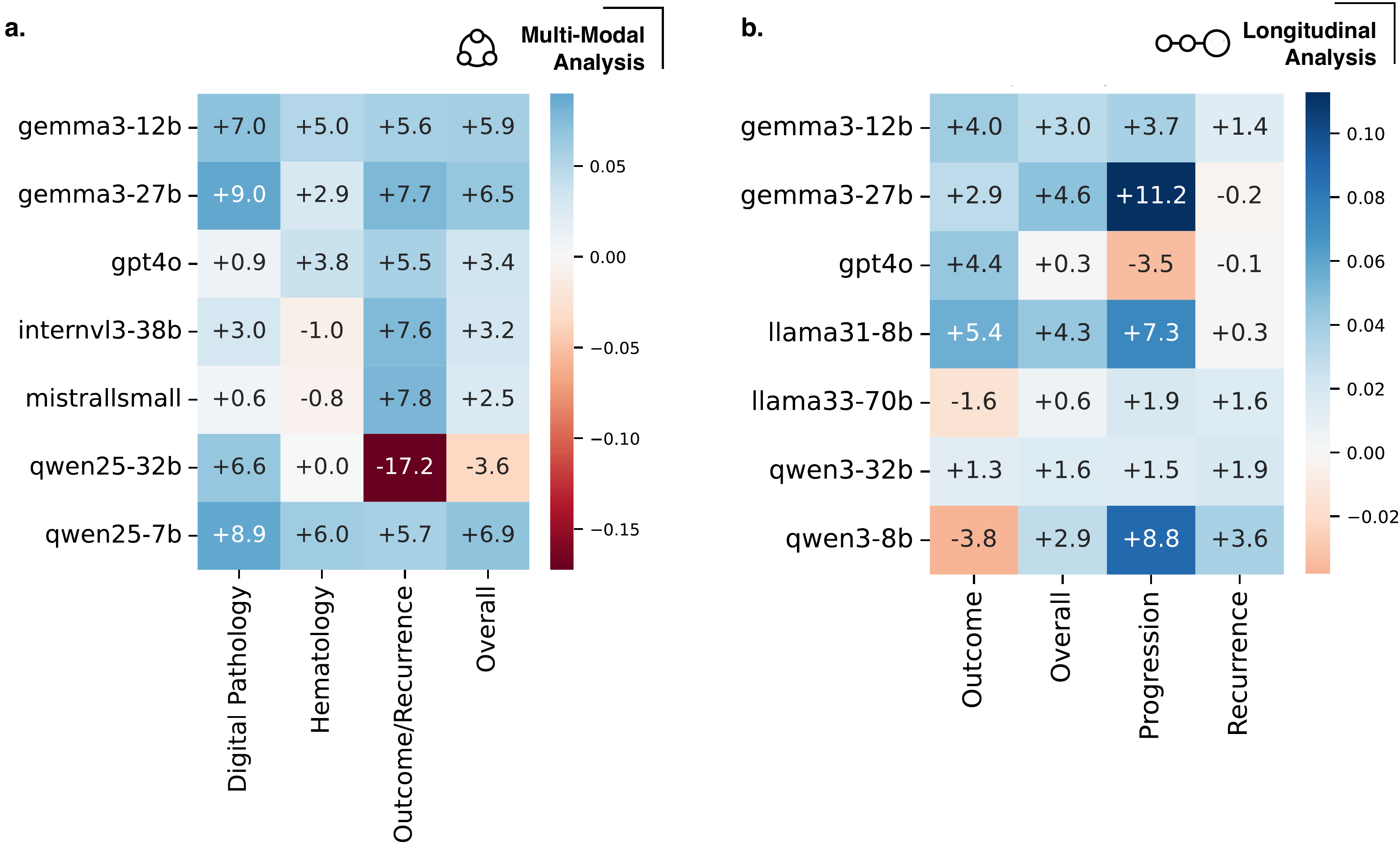}
    \caption{\footnotesize Effect (difference) of tool use on bootstrap accuracy for \textbf{a.} \our-Multimodal and \textbf{b.} \our-Longitudinal.}
    \label{fig:delta_bootstrap_mean}
\end{figure}

Figure~\ref{fig:pc_patients_num_files} shows the distribution of number of files accessed per patient. Stronger models tend to access more files per case, reflecting greater use of available modalities, especially in multi-modal settings. This especially supports the analysis of the Example \ref{fig:example_3}, where models benefit from requesting again detached files in the same conversation. 

Figure~\ref{fig:delta_bootstrap_mean} shows the difference in bootstrap accuracies \withtools\ minus without tools. Smaller models tend to benefit more from tool use (e.g. \qwenvl{7}, \llama{8}, \gemma{12}, \gemma{27}, \qwenvl{32}, \mistralsmall{24}) especially in Digital Pathology, Outcome/Recurrence, and Progression tasks. Notice that \mistralsmall{24} does not benefit that much from tool use in Digital Pathology (+0.6), which could be due to strong vision capabilities. Another notable outlier is \qwenvl{32} which significantly decreases performance with tool use in Outcome/Recurrence tasks. 

\section{Other Prompts}

System prompt used at the beginning of each conversation:
\begin{figure}[H]
    \centering
    \includegraphics[width=0.99\linewidth]{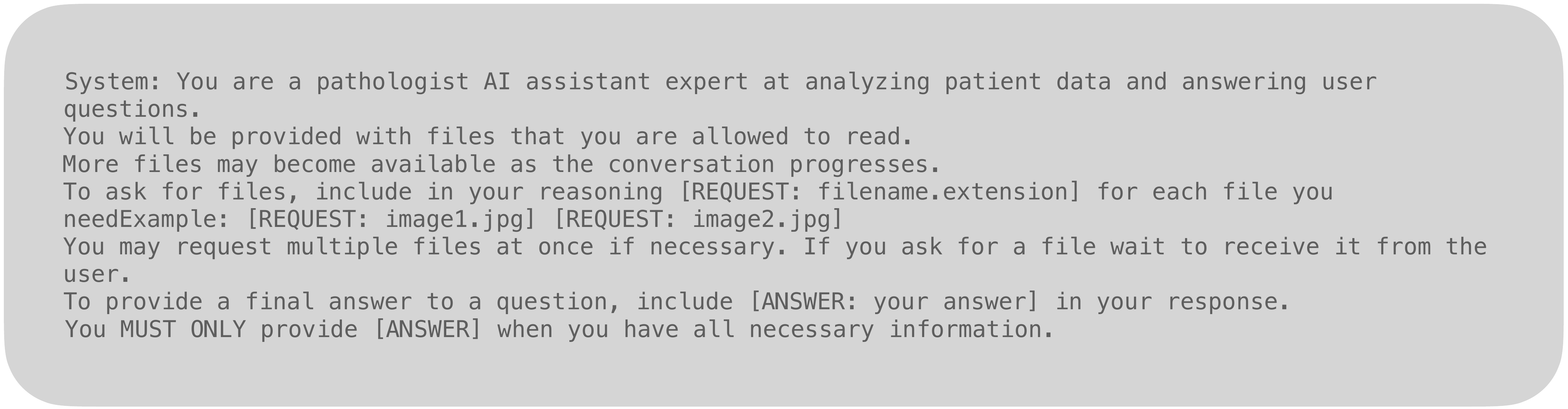}
    \label{fig:sysprompt}
\end{figure}

System prompt when tool calling is enabled:
\begin{figure}[H]
    \centering
    \includegraphics[width=0.99\linewidth]{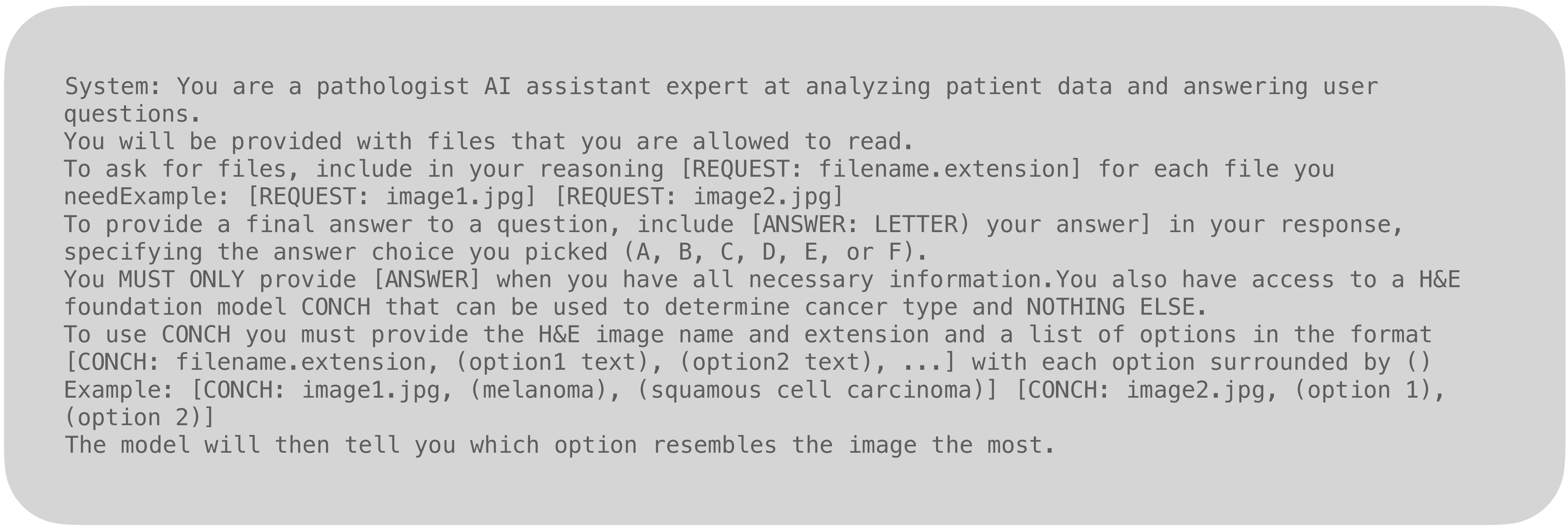}
    \label{fig:sysprompt_tools}
\end{figure}

\clearpage

\end{document}